\documentclass[acmsmall]{acmart}
\usepackage{multirow}
\usepackage{bbold}
\usepackage{amsmath}
\usepackage{url}
\usepackage{subcaption}
\usepackage{pifont} % http://ctan.org/pkg/pifont
\newcommand{\cmark}{\text{\ding{51}}}
\newcommand{\xmark}{\text{\ding{55}}}
\AtBeginDocument{%
  \providecommand\BibTeX{{%
    \normalfont B\kern-0.5em{\scshape i\kern-0.25em b}\kern-0.8em\TeX}}}

\setcopyright{acmcopyright}
\copyrightyear{2022}
\acmYear{2022}
\acmDOI{XXXXXXX.XXXXXXX}

\acmJournal{JACM}
\acmVolume{}
\acmNumber{}
\acmArticle{}
\acmMonth{6}

\begin{document}

%%
%% The "title" command has an optional parameter,
%% allowing the author to define a "short title" to be used in page headers.

%\title{Open-book End-to-End Task-Oriented Dialog (Opera)}

\title{OPERA: Harmonizing Task-Oriented Dialogs and Information Seeking Experience}

%%
%% The "author" command and its associated commands are used to define
%% the authors and their affiliations.
%% Of note is the shared affiliation of the first two authors, and the
%% "authornote" and "authornotemark" commands
%% used to denote shared contribution to the research.
\author{Miaoran Li}
%\authornote{Both authors contributed equally to this research.}
\email{limr@iastate.edu}
%\orcid{1234-5678-9012}

%\author{author2}
%\authornotemark[1]
%\email{email}
\affiliation{%
  \institution{Iowa State University}
%   \streetaddress{P.O. Box 1212}
%   \city{Dublin}
%   \state{Ohio}
   \country{USA}
%   \postcode{43017-6221}
}

\author{Baolin Peng}
\affiliation{%
  \institution{Microsoft Research}
%   \streetaddress{1 Th{\o}rv{\"a}ld Circle}
%   \city{Hekla}
  \country{USA}
}
\email{baolin.peng@microsoft.com}

\author{Jianfeng Gao}
\affiliation{%
  \institution{Micorosoft Research}
%   \city{Rocquencourt}
   \country{USA}
}
\email{jfgao@microsoft.com}

\author{Zhu Zhang}
\affiliation{%
 \institution{Iowa State University}
%  \streetaddress{Rono-Hills}
%  \city{Doimukh}
%  \state{Arunachal Pradesh}
  \country{USA}
}
\email{zhuzhang@iastate.edu}

%%
%% By default, the full list of authors will be used in the page
%% headers. Often, this list is too long, and will overlap
%% other information printed in the page headers. This command allows
%% the author to define a more concise list
%% of authors' names for this purpose.
\renewcommand{\shortauthors}{Li, et al.}

%%
%% The abstract is a short summary of the work to be presented in the
%% article.

\begin{abstract}
  Existing studies in conversational AI mostly treat task-oriented dialog (TOD) and question answering (QA) as separate tasks. Towards the goal of constructing a conversational agent that can complete user tasks and support information seeking, it is important to build a system that handles both TOD and QA with access to various external knowledge. In this work, we propose a new task, Open-Book TOD (OB-TOD), which combines TOD with QA task and expand external knowledge sources to include both explicit knowledge sources (e.g., the Web) and implicit knowledge sources (e.g., pre-trained language models). We create a new dataset  OB-MultiWOZ, where we enrich TOD sessions with QA-like information seeking experience grounded on external knowledge. We propose a unified model OPERA (\underline{Op}en-book \underline{E}nd-to-end Task-o\underline{r}iented Di\underline{a}log) which can appropriately access explicit and implicit external knowledge to tackle the defined task. Experimental results demonstrate OPERA's superior performance compared to closed-book baselines and illustrate the value of both knowledge types~\footnote{The dataset and code will be released to the community.}.
  %the proposed model has strong ability to handle the new task with access to explicit and implicit external knowledge.
  
\end{abstract}

%%
%% The code below is generated by the tool at http://dl.acm.org/ccs.cfm.
%% Please copy and paste the code instead of the example below.
%%


%%
%% Keywords. The author(s) should pick words that accurately describe
%% the work being presented. Separate the keywords with commas.

%%
%% This command processes the author and affiliation and title
%% information and builds the first part of the formatted document.
\maketitle
\def\multiwoz{MultiWOZ}
\def\multiwozqa{OB-MultiWOZ}
\def\baseline{T5 \texttt{(T)}}
\def\closedbook{T5 \texttt{(T + Q)}}
\def\explicit{T5 \texttt{(T + Q)} \textit{w/} EK}
\def\implicitres{\textsc{OPERA}-\texttt{GPT3PM}}
\def\implicitk{\textsc{OPERA}-\texttt{GPT3KB}}
\def\model{\textsc{OPERA}}
\def\task{OB-TOD}

\section{Introduction}
Constructing conversational AI is a task full of challenges and has recently received extensive attention in the natural language processing (NLP) and information retrieval (IR) communities. Specifically, \emph{question answering (QA)} systems and \emph{task-oriented dialog (TOD)} systems are two important categories of conversational agents in practice \cite{Gao18}.

QA systems aim at answering natural language questions by leveraging knowledge from large-scale data sources. \cite{Chen17} proposes to use Wikipedia as an external knowledge source to tackle open-domain QA tasks and introduces a retriever-reader framework, where TF-IDF based retriever is used to find relevant documents from a large corpus, and document reader is designed to obtain the answer from retrieved passages. Later works mainly attempt to improve the performance of the QA task by improving retrieval techniques \cite{Lee19,Lewis20} or expanding the coverage of external knowledge, e.g., incorporating knowledge from heterogeneous sources \cite{Talmor21, Hannan20, Chen21}. As pre-trained language models \cite{Radford19, Brown20,Raffel20} gain popularity, some researchers also start exploring how to utilize implicit knowledge in these models to tackle QA tasks \cite{Roberts20}.

TOD systems, with different goals from QA systems, are expected to complete various user tasks (e.g., booking tickets and finding restaurants) in a conversational manner \cite{Budzianowski18}. Though current TOD systems can complete basic user tasks, they may fail to provide more information required in follow-up questions not contained in pre-defined databases. \cite{Kim20} attempts to enrich TODs with external knowledge and proposes a more challenging TOD setting that requires models to first detect knowledge-seeking turns, then select relevant knowledge snippets from collected FAQs, and finally generate responses. Recent works \cite{Gao21, GaoS22} propose to use extended belief tracking to better manage structured knowledge in pre-defined databases and unstructured knowledge from collected snippets.

Despite numerous efforts to create efficient and high-quality QA systems and TOD systems, single-task systems are still far from intelligent conversational agents as users' information needs in single-task problem settings are well-specified, and the system is not required to be aware of what task it should perform. Human-level conversations usually fuse various information seeking tasks, and therefore conversational agents are expected to know what task they are handling and take appropriate actions. In addition, current studies that attempt to incorporate external knowledge to help knowledge-seeking in TODs only consider limited knowledge sources, e.g., collected FAQ snippets \cite{Kim20}, and fail to include up-to-date knowledge.

\begin{figure}[htp]
  \centering
  \includegraphics[width=\linewidth]{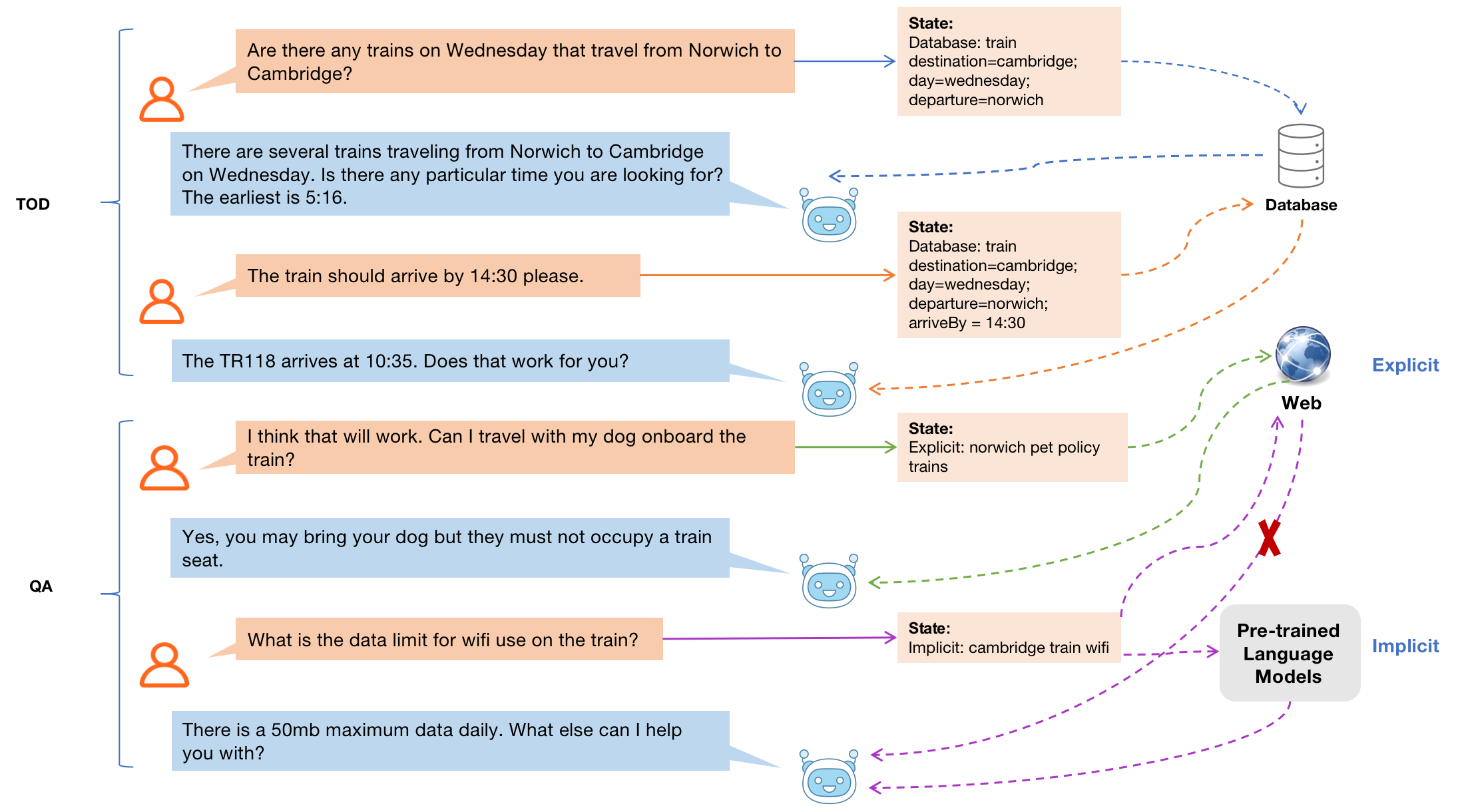}
  \caption{Example of defined task fusing TOD modeling and open-domain QA. The first two turns are similar to traditional TOD while the last two turns are open-domain QA with access to external implicit and explicit knowledge sources. The information required in the last turn can not be found on the Web (explicit knowledge source) and implicit knowledge from pre-trained language models might be helpful.
  }
  \label{fig:introexample}
\end{figure}

Figure~\ref{fig:introexample} shows an example dialog involving both TOD and open-domain QA tasks. The first two turns are similar to traditional TOD tasks requiring information to be found in pre-defined databases. The last two turns, however, need help from external information sources. The system is expected to detect external knowledge seeking turns and consult appropriate knowledge sources. \cite{Nehring21} combines TOD  and open-domain QA using a modular framework, which can serve as a personal assistant and answer questions from Wikipedia. Though it proposes a possible solution to the fused task, it may not be able to handle real-world dialogs with information seeking turns since not all information can be found explicitly. For instance, in the last turn of the example dialog in Figure~\ref{fig:introexample}, the user asks about the limitation of WiFi usage on a train. The system realizes that such information is not included in the pre-defined databases and tries to retrieve an answer from Wikipedia, which fails. Traditional methods tend to generate meaningless answers such as "I do not know", which implies information from databases and the Web is not enough to satisfy the requirements of human-like information seeking.

To address research gaps in fused tasks and insufficient knowledge sources, we propose a more challenging and meaningful dialog modeling task, Open-Book TOD (\task), which not only fuses TOD and QA tasks to better mimic human-level conversations with information-seeking behavior but also generalizes possible external knowledge sources to include the Web and pre-trained language models. We create a new dataset \multiwozqa{} by leveraging existing datasets \cite{Budzianowski18, Kim20}. Inspired by studies on implicit knowledge capacity of pre-trained language models \cite{Petroni19,Roberts20,Wang21}, we present a strategy to integrate implicit knowledge from pre-trained language models with commonly used knowledge sources (e.g., databases and the Web) to tackle more information seeking turns in dialogs and propose an end-to-end system base on previous work \cite{Peng21} that can handle QA turns that require external knowledge in the process of TOD. The system predicts a state based on dialog history and uses the predicted state to select an appropriate knowledge source for information retrieval, whose retrieval results are finally used to generate responses. Experimental results indicate that models with the ability to utilize both external implicit and explicit knowledge outperform models without access to external knowledge or only consulting explicit knowledge sources.

%Our contributions are summarized as below:
Specifically, we expect to make the following contributions:
\begin{itemize}
    \item We formulate a general fused task of modeling both TOD and QA with access to explicit and implicit external knowledge sources, \task, to better mimic the information-seeking process in human-like conversations.
    \item Based on existing datasets, we construct a new dataset \multiwozqa{} for \task. Dialog sessions in the new dataset contain QA turns asking for external information in the process of user task completion and fit real-world conversations better.
    \item A unified model \model{} with the ability to appropriately access explicit and implicit knowledge is constructed to address the new task, which uses a predicted state to handle knowledge source selection and information retrieval, and is trained in an end-to-end manner to jointly optimize TOD and QA modeling. Experimental results show that \model{} has a strong ability to tackle the fused task.
\end{itemize}

\section{Related Work}
\subsection{Dialog Systems}
Generally, dialog systems can be categorized into two classes based on whether it is aimed to accomplish some task proposed by a human user \cite{McTear20}. TOD systems are designed to complete certain tasks by interacting with users, while \emph{open-domain dialog (ODD)} systems are expected to engage in casual conversations.% as the way human interact.  
\subsubsection{TOD Systems}
We have seen progress in methodology for building TOD systems in recent years, from modularized modeling to end-to-end modeling. \cite{Wen17} introduces end-to-end trainable neural networks to TOD modeling. Compared to modular frameworks, which optimize different modules separately, end-to-end training can jointly optimize the model on the full task and lead to better overall performance for task completion. Recently, many researchers have started leveraging pre-trained language models for end-to-end TOD modeling. \cite{Hosseini20} models TOD as a single sequence prediction problem leveraging on GPT-2 \cite{Radford19} for dialogue response generation. \cite{Peng21} applies task-grounded pre-training over external dialogue corpora to further improve performance.

In addition to training strategies, many works focus on expanding knowledge coverage to better handle user requests. Extending traditional TOD where systems can only access structured databases, unstructured knowledge-grounded task-oriented conversational modeling task \cite{Kim20} requires models to decide whether or not to trigger external knowledge access based on dialog history and it can be split into three subtasks: knowledge-seeking turn detection, knowledge selection, and knowledge-grounded generation. Many previous works on this track focus on improving knowledge selection strategy \cite{Thulke21,Jin21, Lotfi21,He21}. Recent works HyKnow \cite{Gao21} and SeKnow \cite{GaoS22} introduce extended belief state to handle TOD grounded on both structured and unstructured knowledge, which is used to perform both database querying and document retrieval. Some efforts \cite{Wu21, Wu20, Wu22} explore how to utilize (heterogeneous) knowledge from multiple sources to make full use of available datasets and, therefore, further improve the applicability of knowledge-enhanced methods. These works shed light on utilizing more knowledge sources to improve knowledge-grounded dialogs. They mainly focus on integrating knowledge collected from different local datasets, while our work utilizes explicit knowledge from the Web and implicit knowledge from pre-trained language models.

Unlike previous works, we do not pay much attention to selecting retrieved knowledge snippets but focus on selecting appropriate knowledge resources. We extend the task of knowledge-seeking turn detection to decide whether to consult external knowledge resources and choose to access the appropriate (implicit or explicit) knowledge source.

\subsubsection{ODD Systems}
Early works \cite{Sordoni15, Vinyals15} without explicit use of knowledge usually suffer from less engaging conversations since they are not grounded in the real world and may fail to provide meaningful responses related to user utterances \cite{Huang20}. To obtain high-quality responses involving more content, many researchers have started exploring how to incorporate dialog systems with external knowledge. \cite{Dinan19} focuses on knowledge-grounded dialog based on a knowledge base comprising Wikipedia articles. To address research difficulties due to scarcity of datasets and limitations of external knowledge sources, Topical-Chat \cite{Gopalakrishnan19} is released, which is a knowledge-grounded conversation dataset built on a knowledge base covering wider topics. The Web is also used as an external knowledge resource \cite{Komeili21} to generate responses with more knowledge and less misinformation. The proposed model for search-engine-augmented generation consists of a query generator and a FiD-style \cite{Izacard21} reader trained separately. \cite{Zhou21} first externalizes implicit commonsense knowledge for grounded response generation by first matching dialogs with an open-source knowledge graph dataset and finally asking models to generate both knowledge and responses given history. In this work, we consider both the Web and pre-trained language models as external knowledge sources.

\subsubsection{Dialog Systems for Fused Task}
To better mimic human-level conversations that usually fuse various information, researchers also start studying fused task settings, such as enhancing TODs with ODDs \cite{Young21, Sun21,Chiu22,Chen22} to make dialogs more natural and engaging. Few works enable TOD systems to answer questions with access to external knowledge. \cite{Nehring21} combines a TOD system and open-domain QA using a modular framework, which can serve as a personal assistant and answer questions from Wikipedia. The system has a module selection component, which selects either DrQA \cite{Chen17} module for open-domain QA or Frankenbot \cite{Bunk20} for TOD completion based on user utterance.

However, formulating TOD and QA independently and simply combining each component prevent information sharing and thus yield suboptimal performance. In this work, we propose \model{}, a unified end-to-end model that jointly models TOD and QA tasks. In addition, existing work either rely on explicit knowledge or implicit for QA while \model{} learns to automatically select which knowledge source is appropriate for each task.

% is not enough to handle real-world dialogs with information seeking turns, since current methods for QA tasks mainly depend on explicit knowledge sources, such as pre-defined database or the web, to obtain knowledge required for answering questions and omit implicit knowledge from pre-trained language models. To cover this gap, we propose a new task setting that requires systems to not only detect turns asking for external information but also choose to access explicit or implicit external knowledge sources. We create a new dataset for defined task based on \multiwoz{} 2.1 \cite{Eric20} and DSTC9 Track1 \cite{Kim20} data collection, consisting of dialogs fusing TOD and QA tasks.
 
\subsection{Open-domain Question Answering}
Open-domain question answering (QA) is a task of finding answers to general-domain questions without specified context by searching a large set of documents (locally or via the Web). \cite{Chen17} tackles the open-domain QA task by accessing Wikipedia as an external knowledge source and constructs DrQA system, which is in the retriever-reader framework. The retriever aims to find relevant passages from a large collection of documents, and the reader component is designed to find an answer from retrieved passages. In early works, the retriever is usually based on TF-IDF or BM25, while the reader is implemented by neural models. In order not to limit QA systems with the recall ceiling of untrainable retrievers, later works start exploring how to implement the retriever by using dense representations \cite{Lee19, Karpukhin20,Lewis20}.

Some researchers attempt to improve performance in QA tasks with a wider range of external knowledge. \cite{Pan19} explores methods for incorporating both open-domain and in-domain external knowledge into a pre-trained model to help with questions that require background knowledge. Recently some researchers have also started focusing on utilizing knowledge from different domains \cite{ChenC21} or heterogeneous modalities \cite{Pramanik21,Li22, Talmor21, Hannan20, Chen21}, such as tables, text, and images.

With the emergence of large-scale pre-trained language models \cite{Radford19, Brown20,Raffel20}, some researchers have started trying to utilize implicit knowledge in pre-trained language models to tackle QA tasks instead of focusing on improving retrieval efficiency and expanding knowledge coverage \cite{Roberts20}. \cite{Wang21} finetunes BART and requires it to answer questions directly with implicit knowledge obtained during the pre-training and finetuning process and finds this is still challenging to answer questions without external knowledge. They suggest one possible solution is to explicitly ask models to retrieve relevant knowledge to improve performance on QA tasks.

In this work, we expect models to consult explicit and implicit knowledge sources appropriately for knowledge acquisition. Compared to previous work that mainly considers answering questions in restricted domains or choosing an option among candidate answers, our task setting is more complex and realistic as the system is expected to respond to multi-domain questions in natural language and learn to access explicit and implicit external knowledge appropriately. We propose to incorporate implicit knowledge from pre-trained language models with explicit knowledge from databases and the Web to better satisfy users' information-seeking requirements.
%by integrating more knowledge.

\subsection{Converational Search and Conversational Question Answering}
Conversational search and conversational question answering (ConvQA) are two subtasks of conversational information retrieval \cite{Gao22,Zamani22}, which is a special information retrieval task in the setting of multi-turn conversations in audio or text. 

There are different definitions of the conversational search task \cite{Anand21,Trippas19,Azzopardi18}. The common property mentioned in various descriptions is that information seeking occurs in a conversational manner, which implies conversations aiming to obtain information may involve multiple exchanges and therefore requires systems to have the capability of handling conversation history and generating an appropriate query to retrieve related passages. 
\cite{Ren18} first defines the task of conversational query understanding, which is a subtask of conversational search that focuses on query understanding in the process of conversational search. The conversational query understanding task is formulated as reformulating a query to include the
necessary context.
\cite{Qu20} introduces OR-QuAC, a benchmark for conversational search, which is created in an open-retrieval conversational question answering setting. Models need to retrieve knowledge from a large collection for answer generation in this setting. Research in this field mainly focuses on retrieving and selecting the most relevant passages. 

Previous work in ConvQA mainly formulates this task in two categories \cite{Gao18}: conversational machine comprehension (CMC) task \cite{Zhu18, Reddy19, Choi18} and sequential knowledge-base question answering (KB-QA). Our work is more related to the CMC task since we only consider unstructured external knowledge. CMC task is usually described as answering questions given a passage (or a set of passages) and the conversation history in the form of question-answer pairs. Most works are conducted on CoQA \cite{Reddy19} and QuAC \cite{Choi18} datasets, where conversations mainly center on the provided passages. Different from previous work on answering complex factual questions, \cite{Qin19} proposes to use machine reading comprehension to generate contentful conversations, which also encompasses chitchat and non-factual responses. A more challenging problem setting is proposed in \cite{Ren21}, in which systems are required to answer a query given query history and a list of retrieved candidate passages from the search engine. Compared to previous work, this new task includes more realistic scenarios where users' questions cannot be answered by a short text span extracted from the given passages.

Our work is similar to a combined task of conversational search and ConvQA, which includes handling dialog history for search query generation and generating responses grounded on knowledge. However, in our setting, the search/QA component is the means, not the ends. %However, our task is different from previous works in ConvQA in the following aspects: 
More specifically, (1) Our main task is to complete task-oriented dialogs in which QA for seeking external information may occur, while ConvQA works mainly focus on answering questions; (2) Instead of already knowing the current turn requires external knowledge, our task requires systems to predict whether it is a knowledge-seeking turn based on dialog history, which sets higher demands on models; (3) Besides widely used external knowledge resources, such as the Web or large knowledge base, we also consider pre-trained LMs with implicit knowledge as possible external knowledge sources; (4) Our task requires models to predict the knowledge source to consult and formulate an appropriate query based on dialog history.
\begin{table}[h]
  \caption{High-level summary of related work. "ConvSearch" is short for conversational search. "PLMs" corresponds to pre-trained language models. "\textit{w/} exp. kn." (or "\textit{w/} imp. kn.") represents QA with access to explicit (or implicit) knowledge. We consider large-scale corpus such as Wikipedia as a Web knowledge source.}
  \label{tab:relatedwork}
  \resizebox{0.8\columnwidth}{!}{
  \begin{tabular}{c|c|c|c|c|c}
    \toprule
    \multirow{2}{*}{Related Work} & \multirow{2}{*}{Task} & \multicolumn{3}{c|}{Knowledge sources} & \multirow{2}{*}{Training method}\\
    \cline{3-5}
    & & Database & Web & PLMs & \\
    \midrule
    \cite{Peng21} & TOD & $\cmark$ & $\xmark$ & $\xmark$ & End-to-end\\
    \cite{Komeili21} & ODD & $\xmark$ & $\cmark$ & $\xmark$ & End-to-end\\
    \cite{Sun21} & TOD + ODD & $\cmark$ & $\xmark$ & $\xmark$ & End-to-end\\
    \cite{Chiu22} & TOD + ODD & $\cmark$ & $\xmark$ & $\xmark$ & Modular\\
    \cite{Nehring21} & TOD + QA w/ exp. kn. & $\cmark$ & $\cmark$ & $\xmark$ & Modular\\
    \midrule
    \cite{Pan19} & QA w/ exp. kn. & $\xmark$ & $\cmark$ & $\xmark$ & End-to-end \\
    \cite{Wang21} & QA w/ imp. kn. & $\xmark$ & $\xmark$ & $\cmark$ & End-to-end\\
    \midrule
    \cite{Qu20} & ConvSearch + ConvQA & $\xmark$ & $\cmark$ & $\xmark$ & End-to-end\\
    \cite{Ren21} & ConvQA & $\xmark$ & $\cmark$ & $\xmark$ & End-to-end\\
    \midrule
    \midrule
    This work & TOD + QA & $\cmark$ & $\cmark$ & $\cmark$ & End-to-end\\
    \bottomrule
  \end{tabular}}
\end{table}
\paragraph{Our work}
Table~\ref{tab:relatedwork} provides a high-level comparison of our work and previous research. There are not many works on the fused task setting of TOD and QA. We propose a new task setting that seamlessly fuses TOD and QA tasks and generalizes the possible knowledge sources to include databases, the Web, and pre-trained language models. We create a new dataset and design an end-to-end model for our task.

\section{Task Formulation}
In this work, we define a new task, Open-book TOD\footnote{\label{fn-open}Please note that the use of ``open-book'' here underscores expanding the TOD agent's capability with access to external knowledge, in contrast to the traditional ``closed-book'' TOD architecture with no expectation of external information seeking. It is different from  ``open book QA''~\cite{Mihaylov18}, where the emphasis is on multi-step reasoning over commonsense knowledge in a  QA setting.} (\task), which fuses two traditional tasks: (1) turn-level TOD modeling, and (2) QA with access to external explicit and implicit knowledge sources. In a traditional TOD modeling task \cite{Budzianowski18}, the system is assumed to access a back-end database to obtain information required by the user. \cite{Kim20} extends TOD task setting to integrate unstructured external knowledge from collected FAQs. 
In this work, we propose a more challenging task extending previous TOD tasks in two aspects: (1) We fuse TOD and QA tasks and require models able to handle QA turns occurring in TODs that ask for external knowledge; (2) We generalize possible external knowledge sources, especially explicit knowledge sources (e.g., the Web) and implicit knowledge sources (e.g., pre-trained language models), to enhance standard TOD systems.

\begin{figure}[htp]
  \centering
  \includegraphics[width=0.8\linewidth]{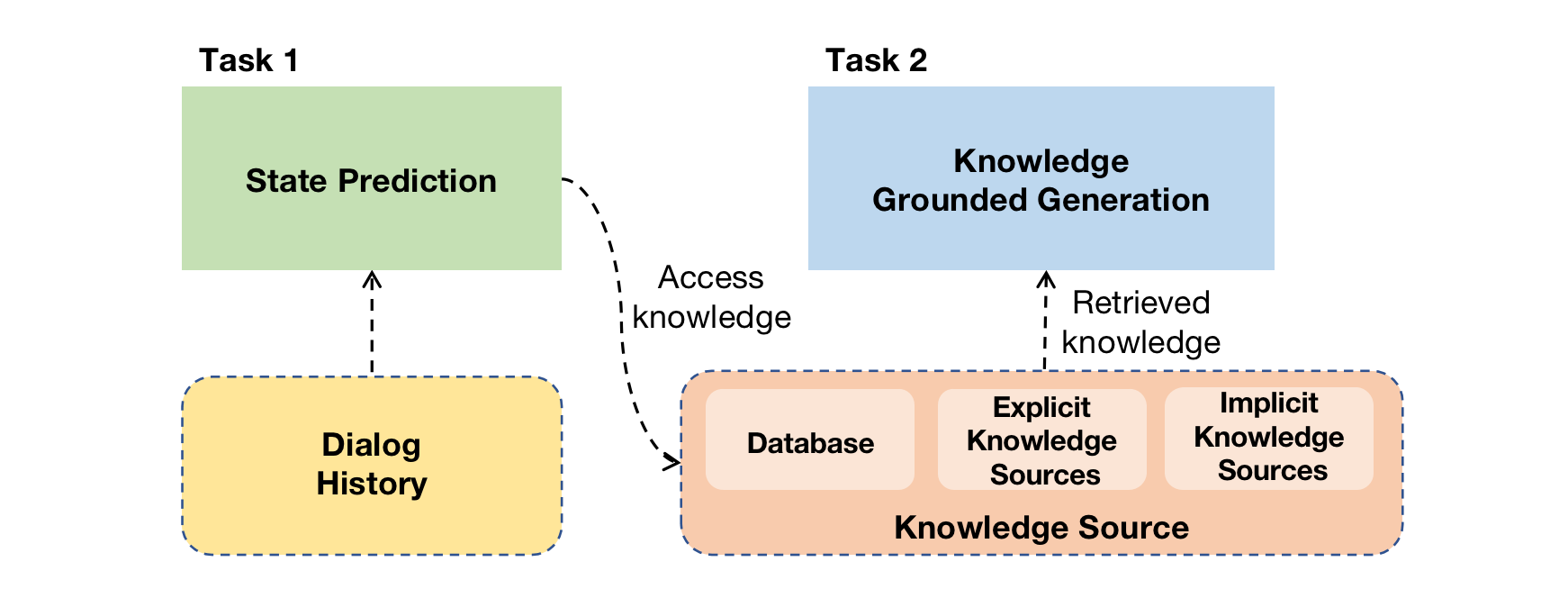}
  \caption{Overview of our task. We split \task{} into two subtasks: (1) state prediction and (2) grounded response generation}
  \label{fig:task}
\end{figure}

Figure~\ref{fig:task} shows the overview of our task. In each dialog turn, the model is required to first predict state $\boldsymbol{s}$ based on dialog history $\boldsymbol{h}$ up to the current turn, which indicates the appropriate knowledge source to access for obtaining knowledge $\boldsymbol{k}$, and finally generate response $\boldsymbol{r}$ grounded on retrieved knowledge. 

We break the full task into two subtasks: state prediction and knowledge-grounded response generation. In $t$-th dialog turn, given dialog history $\boldsymbol h = \{\boldsymbol u_{t-k}, \boldsymbol r_{t-k},..., \boldsymbol u_t\}$, where $\boldsymbol u_i$ and $\boldsymbol r_i$ represent user utterance and system response in the $i$-th turn, respectively, and $k$ is the history window size, the model predicts state $\boldsymbol s$ informing which knowledge source would be accessed and the query used to obtain knowledge from the selected knowledge source, formulated as
\begin{equation}
    \boldsymbol{s} = f_s(\boldsymbol{h})
\end{equation}
If external knowledge is not required, a database state is obtained from the pre-defined database. Otherwise, the model chooses to retrieve either explicit knowledge or implicit knowledge. Database state in plain text or retrieved external knowledge is regarded as knowledge $\boldsymbol{k}$ for response generation. 
We assume that the process of knowledge retrieval is deterministic given the state $\boldsymbol{s}$ because the database lookup process is deterministic, and retrieval results of the Web or pre-trained language models tend to be unchanged.
Finally, the model generate response $\boldsymbol{r}$ based on dialog history $\boldsymbol{h}$, predicted state $\boldsymbol{s}$ and retrieved knowledge $\boldsymbol{k}$:
\begin{equation}
    \boldsymbol{r} = f_r(\boldsymbol{h},\boldsymbol{s},\boldsymbol{k})
\end{equation}

%\section{\multiwozqa}
\section{Constructing an OB-TOD Dataset}

Since exiting TOD or QA datasets do not embody the tight integrative nature of the OB-TOD task, we build a new dataset \multiwozqa{} to facilitate our work and future efforts by the community.

\subsection{Dataset Construction}
%In order to train and evaluate systems that can handle both TOD and QA by accessing appropriate knowledge sources, 
We construct \multiwozqa{} by enriching TOD sessions with QA-like information seeking experience. Specifically, QA turns involving external knowledge are inserted in TOD sessions by leveraging the original \multiwoz{} 2.1~\cite{Eric20} and DSTC9 Track1 dataset \cite{Kim20}. In DSTC9 Track1 data collection \cite{Kim20}, given a sampled dialog from \multiwoz{} 2.1 and a goal subject from external knowledge categories, crowd workers are asked to first identify a suitable position to insert a new knowledge-seeking turn, then write a user utterance at the selected position, and finally generate a system response given relevant knowledge snippets collected from FAQs. Workers are not required to retrieve external knowledge and only need to convert provided information to system utterance. In contrast, we expect workers to retrieve knowledge from the Web and answer user questions based on selected knowledge from retrieval results. We first collect search queries, explicit knowledge, and responses on the Amazon Mechanical Turk, then augment implicit knowledge using GPT-3.

\begin{figure}[htp]
     \centering
     \begin{subfigure}[b]{0.3\textwidth}
         \centering
         \includegraphics[width=\textwidth]{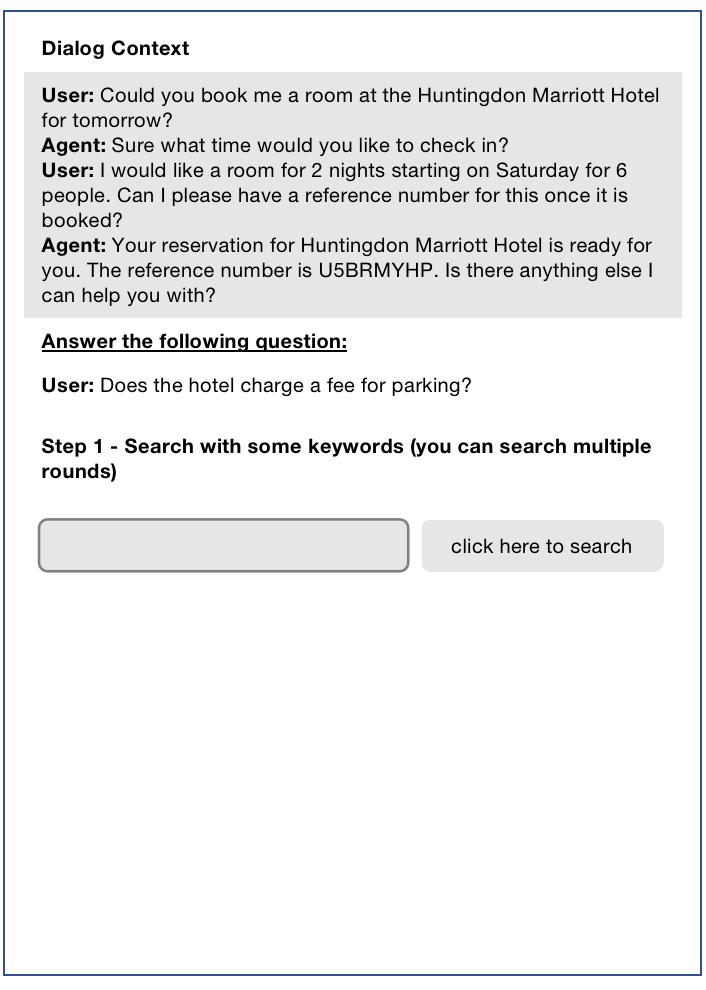}
         \caption{Query Generation}
         \label{fig:query}
     \end{subfigure}
     \hfill
     \begin{subfigure}[b]{0.3\textwidth}
         \centering
         \includegraphics[width=\textwidth]{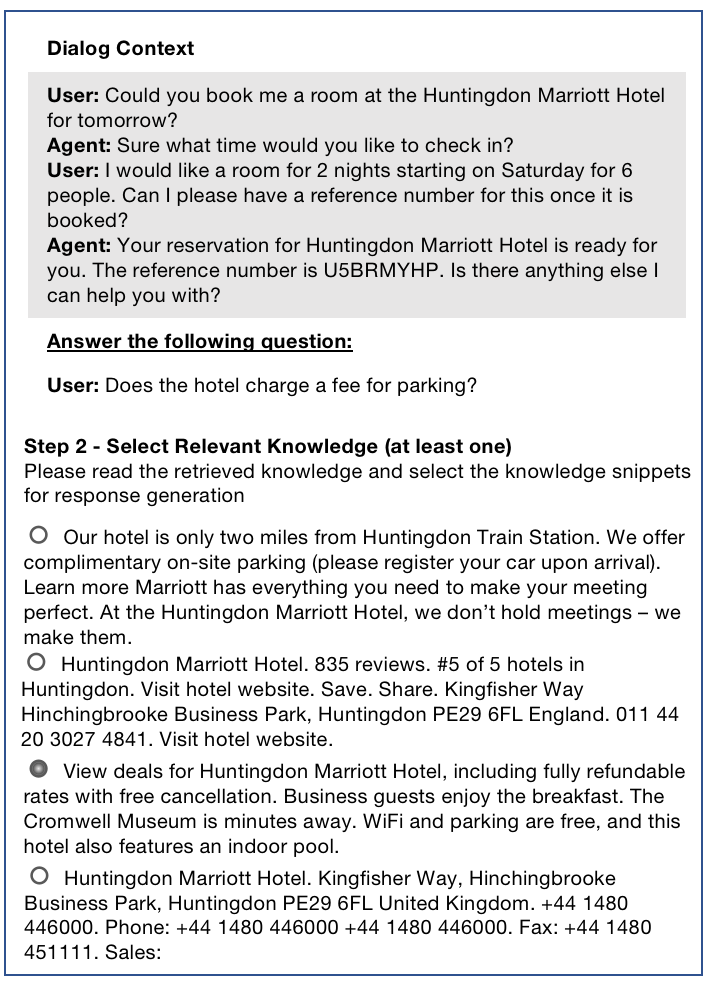}
         \caption{Knowledge Selection}
         \label{fig:select}
     \end{subfigure}
     \hfill
     \begin{subfigure}[b]{0.3\textwidth}
         \centering
         \includegraphics[width=\textwidth]{figures/data_collection1.png}
         \caption{Response Generation}
         \label{fig:response}
     \end{subfigure}
        \caption{The pipeline of data collection. Given dialog context and user utterance, a worker was asked to (a) generate an appropriate query for the user question, (b) select relevant knowledge snippets from retrieval results using the generated query, and (c) write a response based on the selection in (b) or mark question unanswerable. After collecting data from crowd workers, we use GPT-3 to generate implicit knowledge for unanswerable questions.}
        \label{fig:workflow}
\end{figure}

\subsubsection{Crowdsourcing}
The crowdsourcing workflow is shown in Figure~\ref{fig:workflow}. Given a TOD with inserted user utterance for knowledge seeking, crowd workers were asked to write down a search query for the user question and use it to obtain relevant information from a search engine (Figure~\ref{fig:query}). After receiving retrieval results, workers needed to identify whether they were useful. If workers found valuable passages in retrieval results, they were required to select those passages (Figure~\ref{fig:select}). If workers found useful knowledge in retrieval results, they were asked to write a response to the user utterance based on selection. If they found that it is hard to answer the user question based on any of the retrieved passages, they should mark the user question as an unanswerable question (Figure~\ref{fig:response}). Inserted system utterances in the DSTC9 Track1 dataset would be used to respond to unanswerable questions.

\subsubsection{Answerable and Unanswerable Questions}
A user question is answerable if workers can find useful information in retrieval results using the generated query. The responses to answerable questions are generated by workers based on selected knowledge. A user question is considered unanswerable if workers think it is hard to answer it based on retrieved passages. In that case, workers should write "not answerable" as the response (Figure~\ref{fig:response}). Inserted system utterances in the original dataset would be used as responses to unanswerable questions. 
\begin{figure}[htp]
  \centering
  \includegraphics[width=0.78\linewidth]{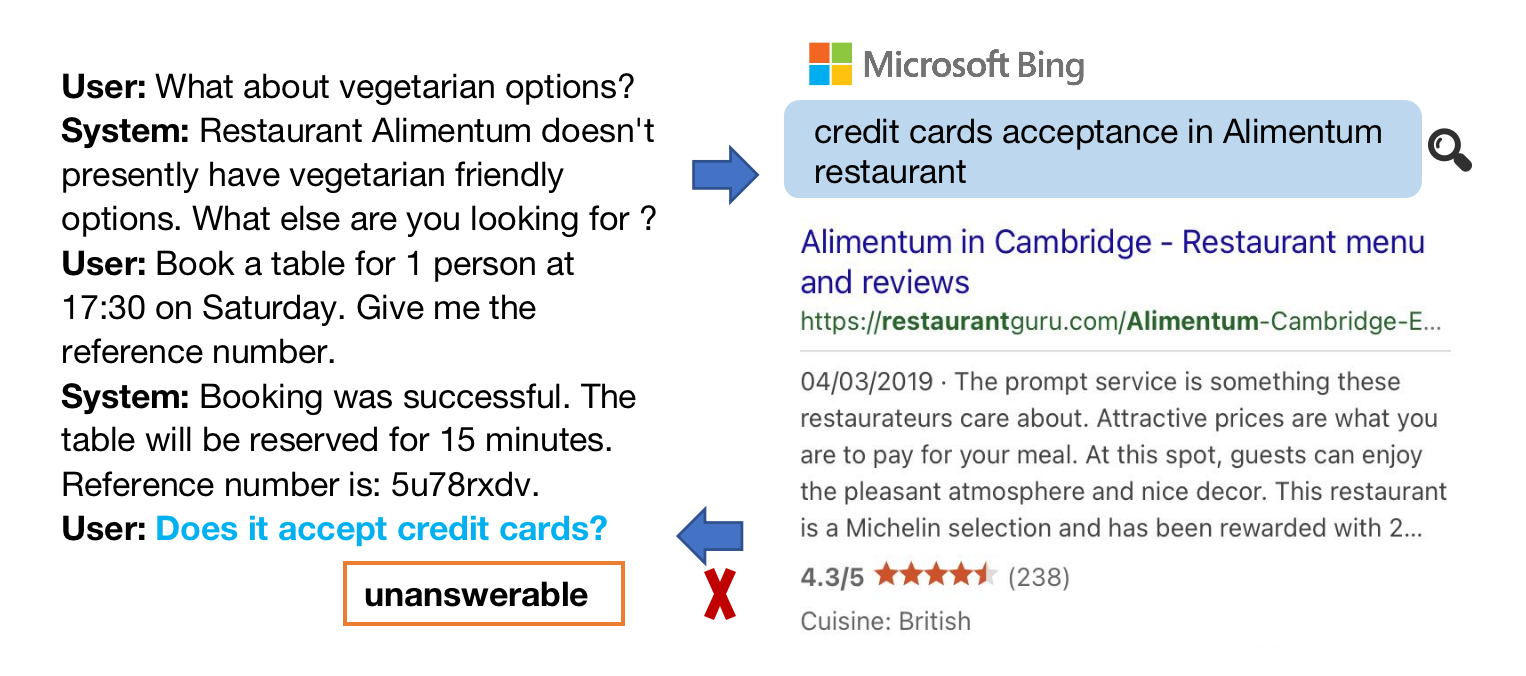}
  \caption{Example of an unanswerable question. Given dialog history and user utterance (left), the search query "credit cards acceptance in Alimentum restaurant" was created (in the blue box) to retrieve information via Bing search API. The crowd worker found none of the retrieved passages could be used to generate a response to the user question and therefore annotated this QA turn as unanswerable.}
  \label{fig:unans}
\end{figure}

Figure~\ref{fig:unans} gives an example of unanswerable question. Dialog history and current user utterance (in blue) are shown on the left side. Crowd worker used the search query "credit cards acceptance in Alimentum restaurant" (in the blue box) to retrieve information via Bing search engine. Top-ranked retrieved passages are listed on the right side. The worker found none of these snippets provided useful information, marking this user utterance as an unanswerable question. 

\subsubsection{Implicit Knowledge Augmentation}
\begin{figure}[htp]
  \centering
  \includegraphics[width=\linewidth]{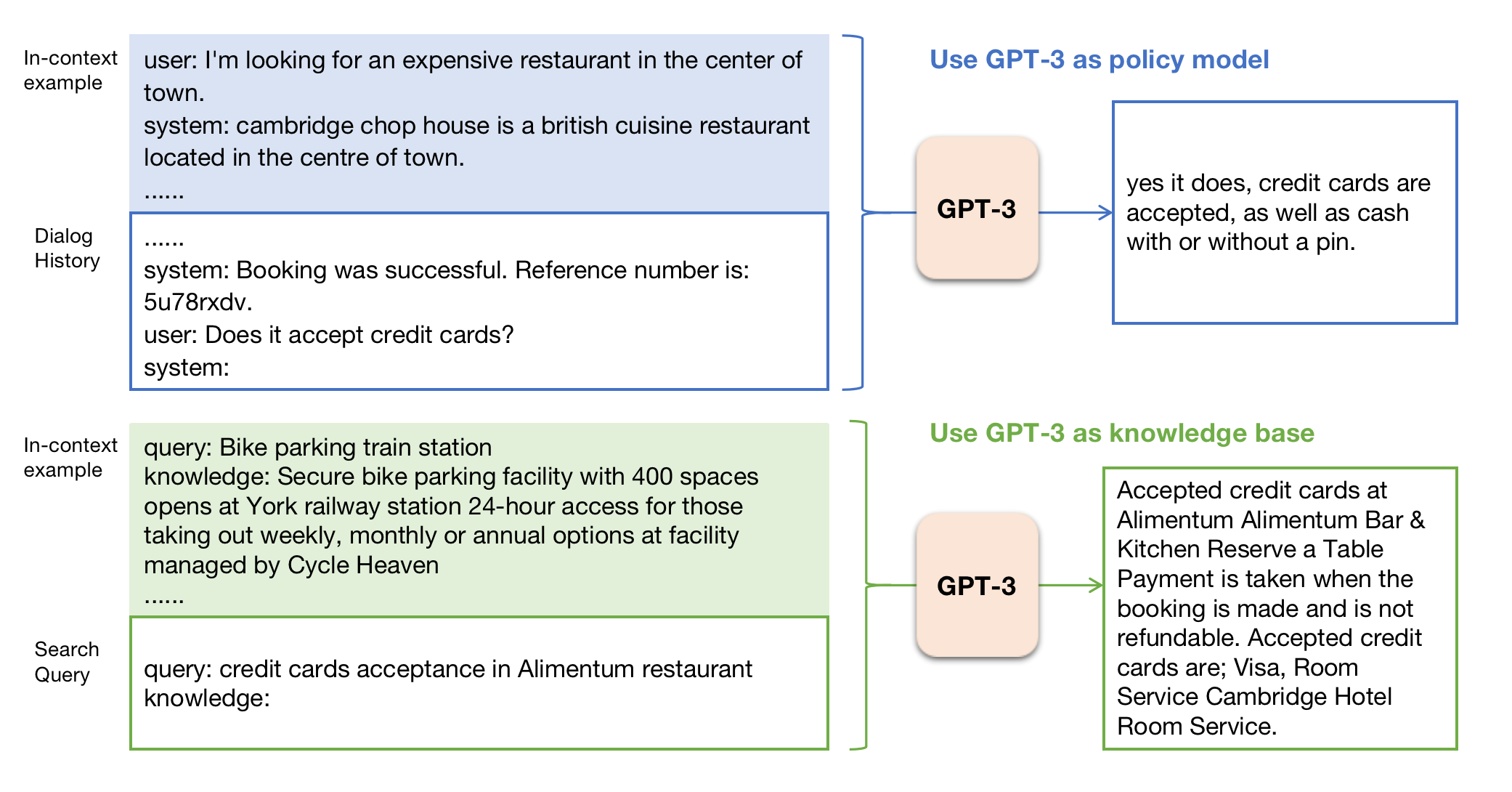}
  \caption{Example of two methods to obtain implicit knowledge using GPT-3. The upper part (in blue) shows an example of using GPT-3 as a policy model to obtain implicit knowledge. The lower part (in green) gives an example of using GPT-3 as a knowledge base for implicit knowledge acquisition.}
  \label{fig:implicit}
\end{figure}

Since unanswerable questions cannot be answered using explicit knowledge, they might benefit from implicit knowledge in pre-train language models. After collecting search queries, selected knowledge (explicit knowledge), user question types (answerable or unanswerable), and responses, we further augmented the dataset by adding implicit knowledge for unanswerable QA turns. We considered two methods to generate implicit knowledge (Figure~\ref{fig:implicit}) by using GPT-3 \cite{Brown20}. GPT-3 is proven to have the ability of in-context learning, which means it can be quickly adapted to new tasks with only a few examples in the inference phrase without fine-tuning.

The first method is to use GPT-3 as a policy model to generate implicit knowledge. We inputted two sampled dialogs as in-context learning examples, concatenated with current dialog history, and the generated system response is used as implicit knowledge. GPT-3 is expected to learn from in-context learning examples and, therefore, generate a reasonable response to the current user utterance. The in-context example was in the form of \texttt{<user utterance>$\setminus$n<system utterance>$\setminus$n$\cdots$ <system utterance>$\setminus$n$\setminus$n}, where \texttt{$\setminus$n} is used as in-dialog separator and \texttt{$\setminus$n$\setminus$n} is used to separate different sample dialogs. The upper part of Figure~\ref{fig:implicit} shows the example of implicit knowledge generation by using GPT-3 as a policy model based on dialog history for the unanswerable question shown in Figure~\ref{fig:unans}.

The other method is to use GPT-3 as a knowledge base to obtain implicit knowledge. We inputted the concatenation of in-context examples and annotated query to GPT-3, where the in-context example was compromised of search queries and selected knowledge of sampled answerable questions in the form  \texttt{<query>$\setminus$n<knowledge>$\setminus$n$\setminus$n}. Using this prompt format, GPT-3 is expected to learn to generate a knowledge snippet similar to explicit knowledge of answerable questions based on a search query. The lower part of Figure~\ref{fig:implicit} shows an example of implicit knowledge generation by using GPT-3 as a knowledge base based on annotated search query for the unanswerable question shown in Figure~\ref{fig:unans}.

\subsection{Statistics}
Table~\ref{tab:statistics} includes basic statistics of the collected dataset.
We used 1573 dialogs with 1705 inserted user utterances from the DSTC9 Track1 dataset for data collection. There are 1366 inserted user questions annotated as answerable questions, while 339 utterances are marked unanswerable.

Statistics of answerable and unanswerable QA turns are shown in Table~\ref{tab:qastatistics}. We do not report the average length of implicit knowledge since we set the average length of explicit knowledge as the target length when generating implicit knowledge.

\begin{table}[htp]
  \caption{Statistics of the data divided into training, validation and test set.}
  \label{tab:statistics}
  \resizebox{0.9\columnwidth}{!}{
  \begin{tabular}{c|cccc}
    \toprule
    Split & \# Dialogs & \# TOD turns & \# Answerable questions & \# Unanswerable questions \\
    \midrule
    Train & 379 & 3094 & 641 & 63\\
    Validation & 41 & 317 & 78 & 7\\
    Test & 782 & 5712 & 647 & 269\\
    \midrule
    Total & 1202 & 9123 & 1366 & 339\\
    \bottomrule
  \end{tabular}
  }
\end{table}

\begin{table}[htp]
  \caption{Statistics of answerable and unanswerable QA turns. The average number of tokens is reported.}
  \label{tab:qastatistics}
  \resizebox{0.73\columnwidth}{!}{
  \begin{tabular}{c|ccc}
    \toprule
    Question Type & Query Length & Knowledge Length & Response Length \\
    \midrule
    Answerable & 33 & 314 & 60\\
    Unanswerable & 37 & \textemdash & 96\\
    \bottomrule
  \end{tabular}}
\end{table}

\section{Model}

\begin{figure}[htp]
  \centering
  \includegraphics[width=0.9\linewidth]{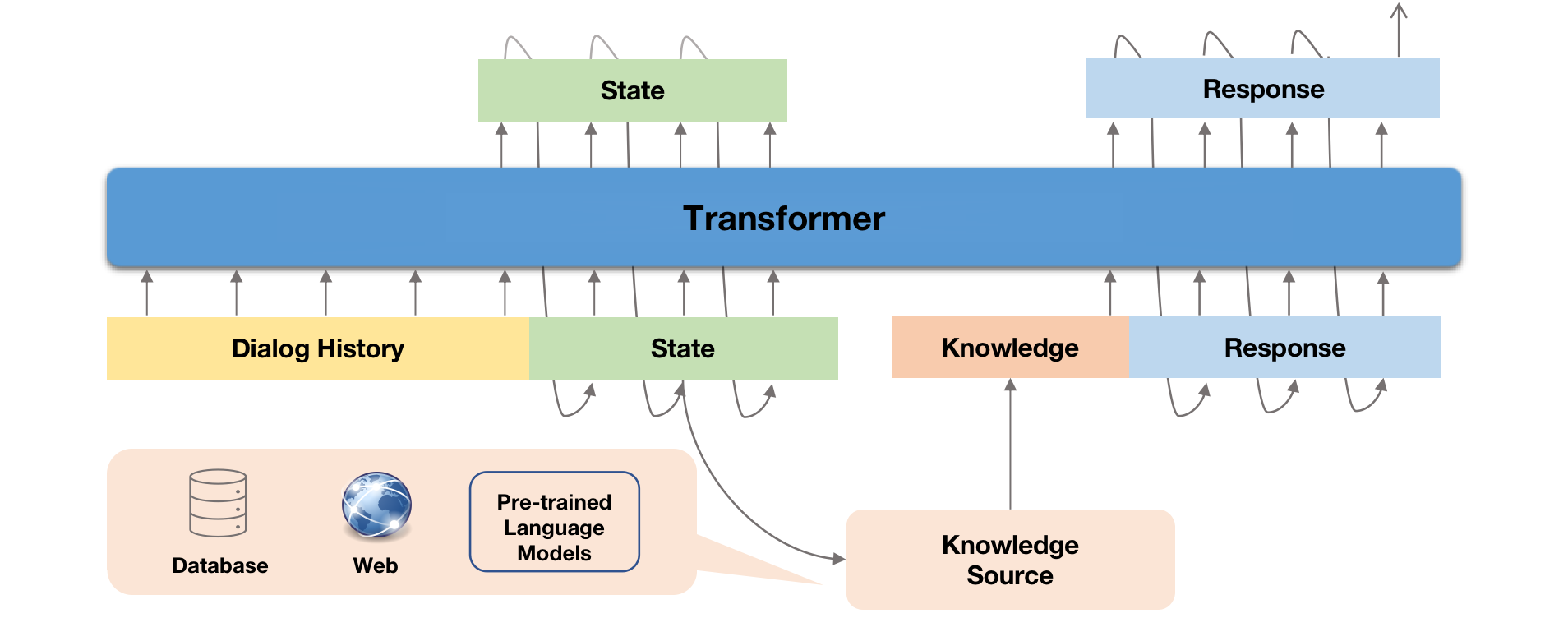}
  \caption{Overall architecture of the OPERA model}
  \label{fig:model}
\end{figure}

We propose a unified model \model{} (\underline{Op}en-book\footnote{Please see footnote \ref{fn-open} regarding the ``open-book'' terminology.} \underline{E}nd-to-end Task-o\underline{r}iented Di\underline{a}log) for the \task{} task. \model{} is novel in two aspects: (1) it seamlessly incorporates QA into end-to-end task-oriented dialogs, and (2) it is equipped with a knowledge source selection mechanism to leverage explicit and implicit knowledge for task completion. The illustration of the proposed model is shown in Figure~\ref{fig:model}. \model{} first predicts states, which track the user's goal, given the dialog history, then queries a knowledge source according to the predicted state for knowledge acquisition, and finally performs response generation grounded on the dialog history, predicted states, and retrieved knowledge.

% We construct a Transformer-based \cite{Vaswani17} model \model{} (\underline{Op}en-book\footnote{Please see footnote \ref{fn-open} regarding the ``open-book'' terminology.} \underline{E}nd-to-end Task-o\underline{r}iented Di\underline{a}log) for the \task{} task. The illustration of proposed model is shown in Figure~\ref{fig:model}. The model first predicts state based on dialog history, then queries a knowledge source according to predicted state to access knowledge, and finally performs response generation grounded on dialog history, predicted state and retrieved knowledge.

\subsection{State Prediction}
State $\boldsymbol{s}$ tracks a user's goal throughout a dialog. In particular, a state $\boldsymbol{s}$ is in the form \texttt{ks: q}, where \texttt{ks} represents the knowledge source to consult, and \texttt{q} stands for query used to acquire knowledge from the predicted source. In this work, we consider three possible knowledge sources: pre-defined database, explicit external knowledge source, and implicit external knowledge source. The query for accessing a pre-defined database is a belief state in traditional TOD modeling, and the query for accessing explicit or implicit external knowledge sources is in the form of a search query.
\begin{equation}
    \boldsymbol{s}=
    \begin{cases} 
      \textit{Database: } \text{belief state}, & \text{model predicts to access database}, \\
      \textit{Explicit: } \text{search query}, &  \text{model predicts to access explicit knowledge source},\\
      \textit{Implicit: } \text{search query}, & \text{model predicts to access implicit knowledge source}.
   \end{cases}
\end{equation} 

\begin{figure}[htp]
     \centering
     \begin{subfigure}[b]{\textwidth}
         \centering
         \includegraphics[width=0.85\textwidth]{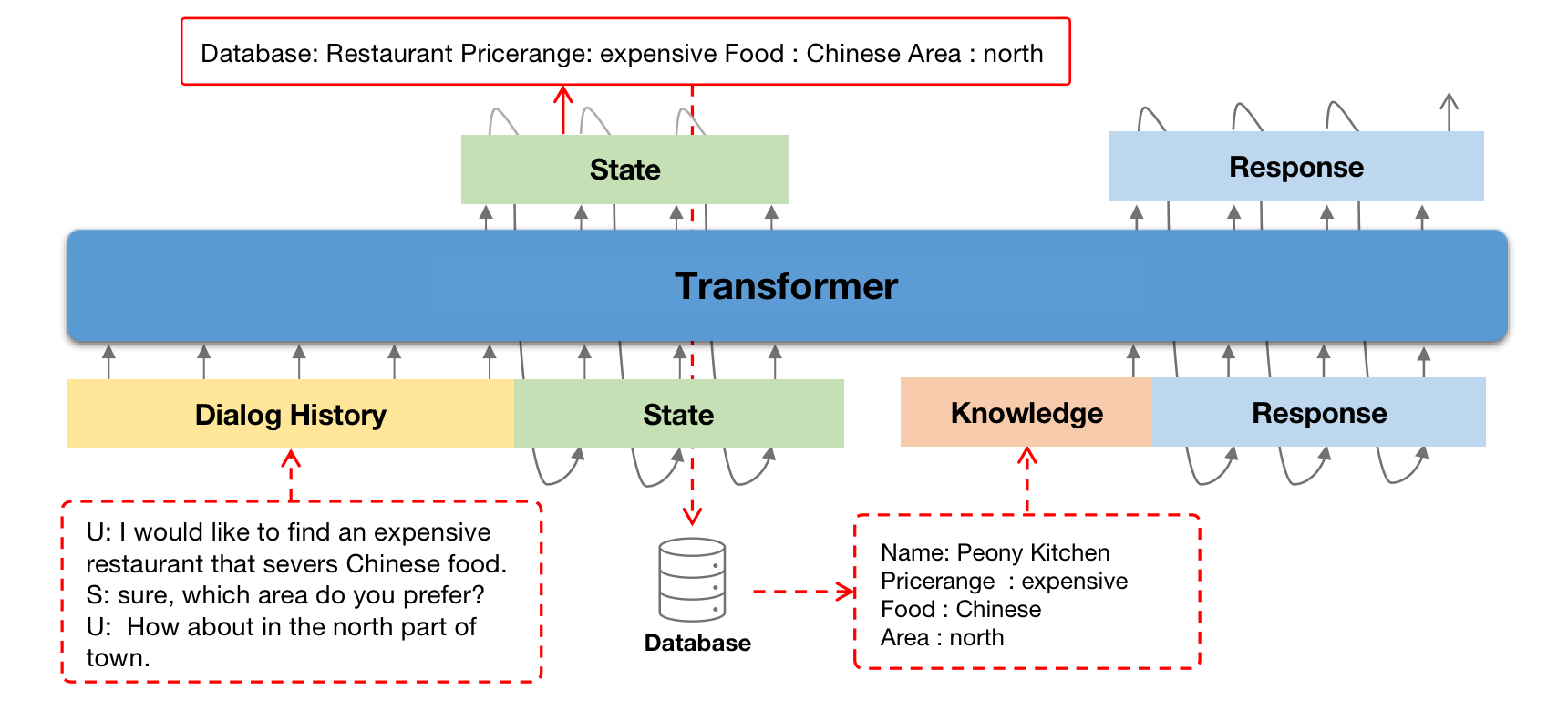}
         \caption{Example of using pre-defined database}
         \label{fig:database_eg}
     \end{subfigure}
     \hfill
     \begin{subfigure}[b]{\textwidth}
         \centering
         \includegraphics[width=0.85\textwidth]{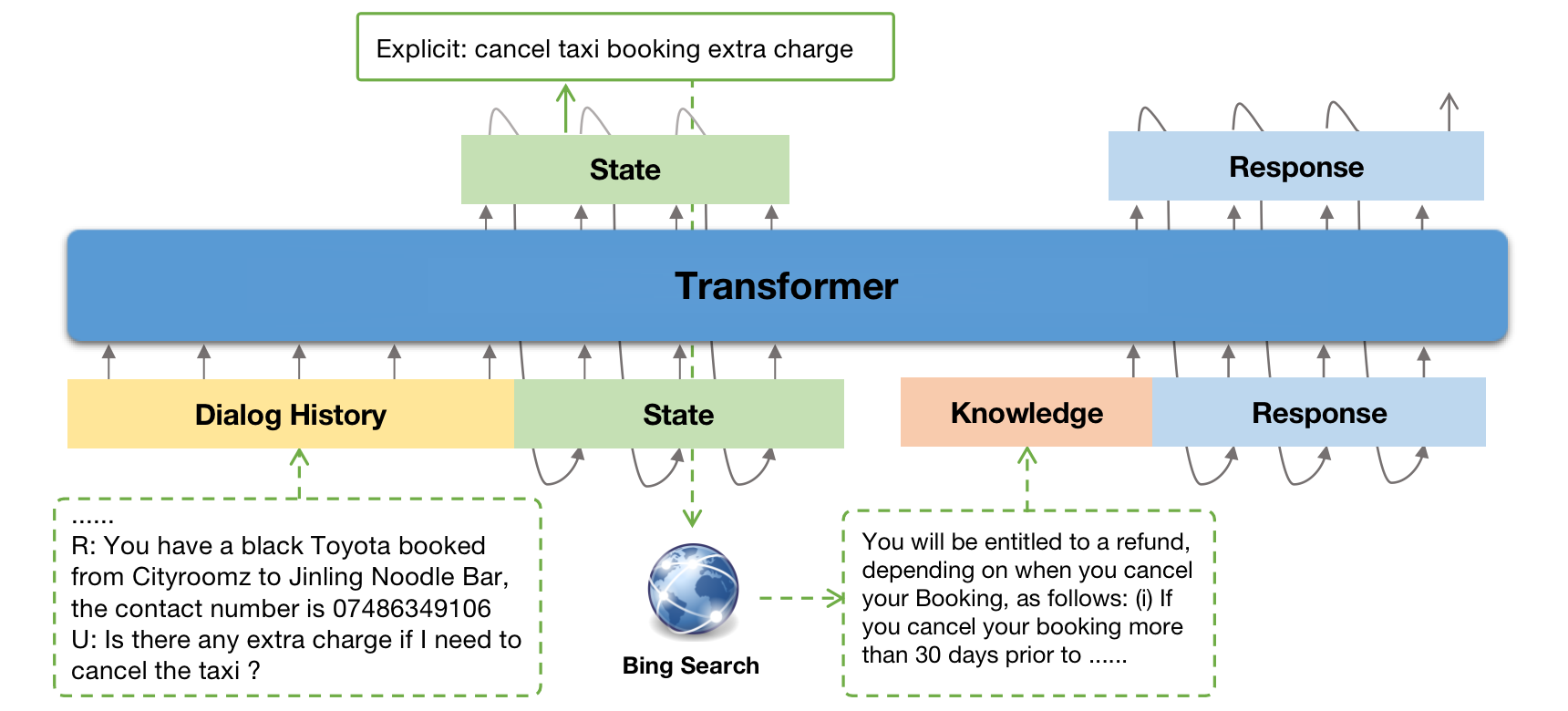}
         \caption{Example of using external explicit knowledge source}
         \label{fig:explicit_eg}
     \end{subfigure}
     \hfill
     \begin{subfigure}[b]{\textwidth}
         \centering
         \includegraphics[width=0.85\textwidth]{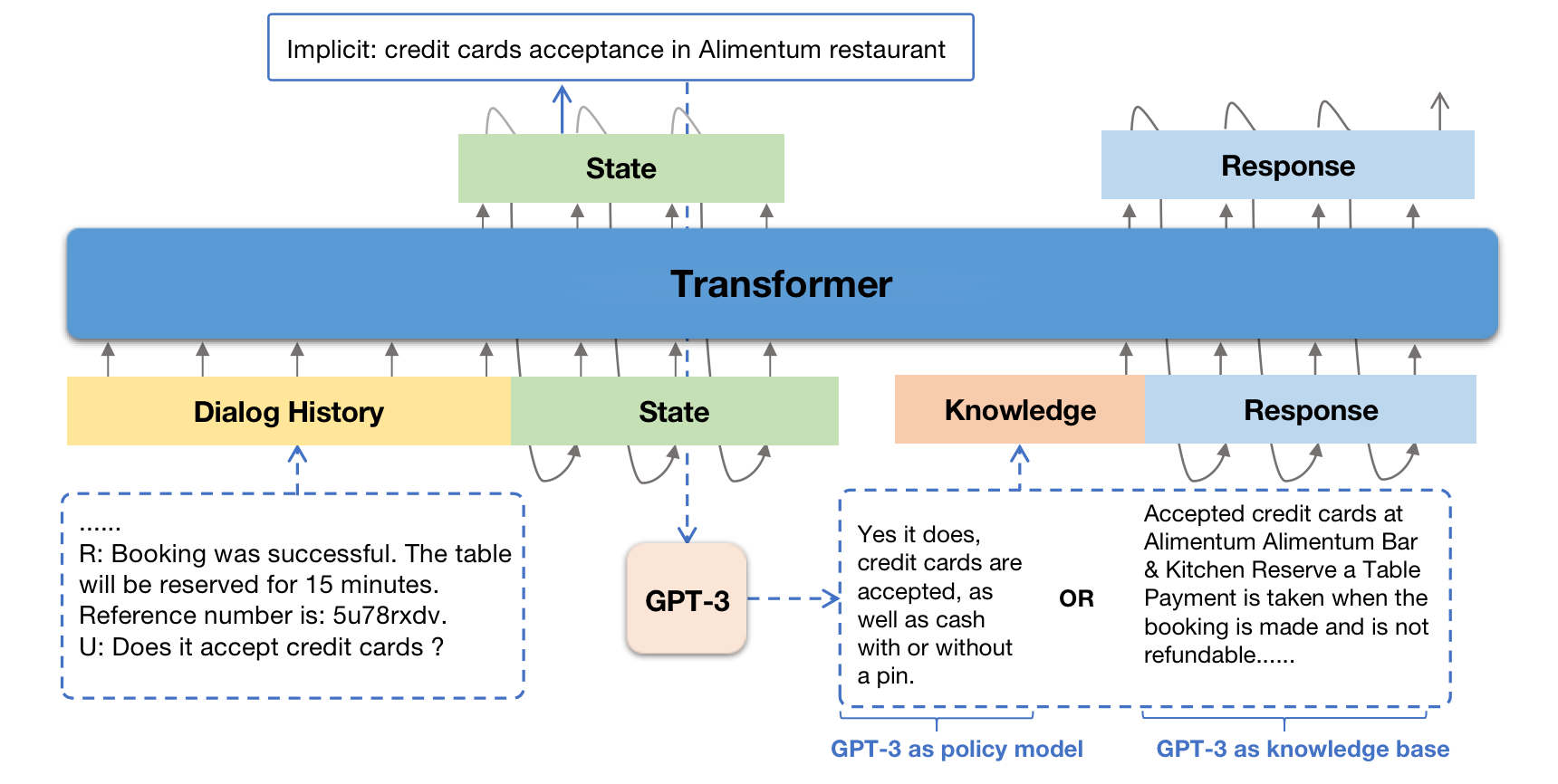}
         \caption{Example of using external implicit knowledge source. Two methods are used to obtain implicit knowledge.}
         \label{fig:implicit_eg}
     \end{subfigure}
        \caption{Examples of the proposed model using different knowledge sources}
        \label{fig:examples}
\end{figure}

Examples of possible states are shown in Figure~\ref{fig:examples}. Suppose the model predicts that external knowledge is not needed to respond current turn (Figure~\ref{fig:database_eg}). In that case, the predicted state indicates that the selected knowledge source is the pre-defined database followed by a belief state, e.g., \textit{Database: restaurant pricerange = expensive food = Chinese area = north}. If the model predicts that the user question is seeking external information, either explicit knowledge source (Figure~\ref{fig:explicit_eg}) or implicit knowledge source (Figure~\ref{fig:implicit_eg}) can be accessed by the predicted search query, e.g., \textit{Explicit: cancel taxi booking extra charge} or \textit{Implicit: credit cards acceptance in Alimentum restaurant}.

Given dialog history $\boldsymbol h = \{\boldsymbol u_{t-k}, \boldsymbol r_{t-k},..., \boldsymbol u_t\}$, where $\boldsymbol u_i$ and $\boldsymbol r_i$ represent user utterance and system response at turn $i$, respectively, and $k$ is the history window size, the training objective of state prediction can be formulated as
\begin{equation} \label{eq:state}
    \mathcal{L}_S = \log p(\boldsymbol{s}\mid \boldsymbol{h}) = \sum_{i=1}^{N_t} \log p_{\boldsymbol{\theta}} (s_i\mid s_{<i},\boldsymbol{h}),
\end{equation}
where $\theta$ represents trainable parameters in the model, $N_t$ is the target length of predicted state sequence, and $s_{<i}$ denotes tokens before index $i$. In the implementation, we add a task-specific prefix \cite{Raffel20} to input to specify which task the model should perform. The input of state prediction is in the form \texttt{State Prediction: <dialog history>}.

\subsection{Knowledge Acquisition}
\subsubsection{Database} 
If \model{} predicts to query the pre-defined database (Figure~\ref{fig:database_eg}) based on predicted state $\boldsymbol{s}$, predicted belief state, which is a list of triplets in the form (\textit{domain, slot\_name, value}) recording values for slots in a particular domain, is used to query the database \cite{Budzianowski18}. A database state that contains records satisfying the conditions of the belief state is returned and used as knowledge for response generation.

\subsubsection{Explicit Knowledge Source}
We use Bing Search API as an explicit knowledge source, which is a black box in the proposed system and is not trainable. It can be easily generalized to other explicit knowledge sources such as Wikidump. Consider the example in Figure~\ref{fig:explicit_eg}, the predicted state $\boldsymbol{s}$ is \textit{Explicit: cancel taxi booking extra charge}. Bing API is triggered, and retrieval results based on query \textit{cancel taxi booking extra charge} are returned.

\subsubsection{Implicit Knowledge Source}
\begin{figure}[h]
  \centering
  \includegraphics[width=0.95\linewidth]{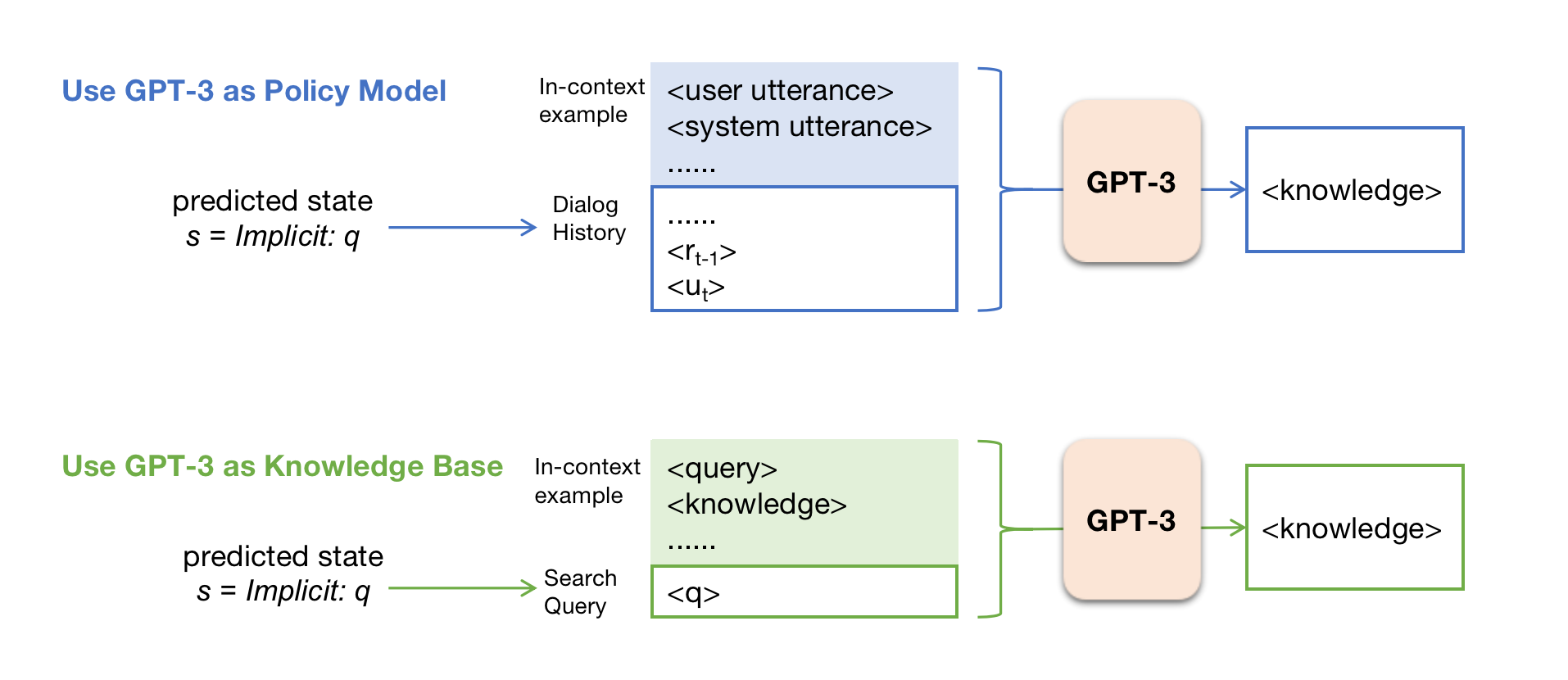}
  \caption{Illustration of two methods to access implicit knowledge source. We use GPT-3 either as a policy model or knowledge base to acquire implicit knowledge.}
  \label{fig:implicit_retrieval}
\end{figure}

In our implementation, we regard GPT-3 \cite{Brown20} as an implicit knowledge source (Figure~\ref{fig:implicit_eg}), which can be replaced by other large-scale pre-trained language models. GPT-3 is proven to have the ability of in-context learning, which means it can be quickly adapted to new tasks with only a few examples in the inference phrase without fine-tuning. As shown in Figure~\ref{fig:implicit_retrieval}, we propose two types of approaches to obtaining implicit knowledge from GPT-3.

\noindent \paragraph{Using GPT-3 as a policy model}
The intuition is to ask GPT-3 to respond to user utterances directly and utilize the generated response as implicit knowledge. Generating response to user questions requires GPT-3 to complete the closed-book QA task \cite{Roberts20}, which involves utilizing implicit knowledge obtained in the pre-training process. The input to GPT-3 is comprised of in-context examples and dialog history. Two example passages are provided for in-context learning. The example is in the form of \texttt{<user utterance>$\setminus$n<system utterance>$\setminus$n$\cdots$ <system utterance>$\setminus$n$\setminus$n}. In-context examples and dialog history are concatenated as input prompt to GPT-3 (upper part of Figure~\ref{fig:implicit_retrieval}), and the generated system utterance is used as implicit knowledge.

\noindent \paragraph{Using GPT-3 as a knowledge base}
The other method to acquire implicit knowledge is to query GPT-3 as the way we request Bing API. The input prompt to GPT-3 consists of in-context examples and the predicted query. The in-context example is compromised of search queries and selected knowledge of sampled answerable QA turns, in the form: \texttt{<query>$\setminus$n<knowledge>$\setminus$n$\setminus$n}. The query in the predicted state is appended to in-context learning examples, and the concatenated string is used as an input prompt to GPT-3 (lower part of Figure~\ref{fig:implicit_retrieval}).

\subsection{Grounded Response Generation}
System response $\boldsymbol{r} = \{r_1,r_2,...,r_{N_r}\}$ with length $N_r$ is generated grounded on dialog history $\boldsymbol{h}$, predicted state $\boldsymbol{s}$ and retrieved knowledge $\boldsymbol{k}$. 
The objective is defined as
\begin{equation} \label{eq:response}
    \mathcal{L}_R = \log p(\boldsymbol{r}\mid \boldsymbol{h},\boldsymbol{s}, \boldsymbol{k}) = \sum_{i=1}^{N_r} \log p_{\boldsymbol{\theta}} (r_i\mid r_{<i},\boldsymbol{h}, \boldsymbol{s}, \boldsymbol{k}).
\end{equation}
% With task-specific prefix for grounded generation, the input is in the form \texttt{Response Generation: <dialog history> <|knowledge|> <knowledge>}, where \texttt{<|knowledge|>} is a special token indicating the beginning of retrieved knowledge.

With task-specific prefix for grounded generation, the input is in the form \texttt{Response Generation: <dialog history> <|knowledge|> <selected knowledge>}, where \texttt{<|knowledge|>} is a special token indicating the beginning of retrieved knowledge.

\subsection{Training Objective of Full Task}

Each training example is represented as:
\begin{equation}
    \boldsymbol x = (\boldsymbol h, \boldsymbol s, \boldsymbol k, \boldsymbol r),
\end{equation}
where $\boldsymbol h = \{\boldsymbol{u_{t-k}}, \boldsymbol{r_{t-k}},...,\boldsymbol{u_t}\}$ is the dialog history consisting of user and system utterance in the last $k-1$ turns and current user utterance, $\boldsymbol s$ is the state including knowledge source to retrieve and the corresponding query, $\boldsymbol k$ is the retrieved knowledge, and $\boldsymbol r$ is the (delexicalized) dialog response. The knowledge $\boldsymbol k$ can be a database state in plain text, a passage in Bing retrieval results, or implicit knowledge from GPT-3. 

Combing learning objectives (\ref{eq:state}) and (\ref{eq:response}) of subtasks, the joint objective of full task is 
\begin{equation}
    \mathcal{L}_{\boldsymbol{\theta}}(\mathcal{D}) = \sum_{i=1}^n(\mathcal{L}_S(\boldsymbol{x_i}) + \mathcal{L}_R(\boldsymbol{x_i})),
\end{equation}
where $\mathcal{D} = \{\boldsymbol{x}_i\}_{i=1}^n$ is training dataset consisting of $n$ training examples. We use a Transformer to parameterize both state prediction and grounded response generation process, and parameters in $\theta$ are updated by maximizing the joint objective.

\section{Experiments}
We perform end-to-end evaluations of the proposed model to answer two questions: (1) How does the proposed model handle \task{} task? (2) How does the proposed model leverage explicit and implicit knowledge to respond to information-seeking questions when completing TODs?

\subsection{Experimental Setup}
\subsubsection{Evaluation Metrics and Datasets}
We consider three evaluation settings: (1) standard TOD completion as described in \cite{Budzianowski18, Eric20}, (2) QA task, and (3) full task involving TOD and QA.

In the TOD evaluation setting, following previous works \cite{Budzianowski18, Eric20} we measure whether the model provides an appropriate entity (\verb|Inform|), e.g., restaurant location or price range, and answers all the request attributes (\verb|Success|), e.g., phone number or postcode, in dialog-level. We use \verb|BLEU| \cite{Papineni02} to measure how fluent the generated responses are compared to human-annotated answers. A combined score \verb|Combine| = (\verb|Inform|+\verb|Success|) $\times$ 0.5 + \verb|BLEU| is computed as an overall measure of generation quality.  

In the QA setting, we measure whether the model predicts correct external knowledge sources to consult (\verb|Accuracy|), generates appropriate search queries (\verb|Query F1|) and succeeds in QA turns (\verb|Success Rate|). We report \verb|BLEU| score of generated answers for QA turns. \verb|Query F1| is the F1 score between a predicted search query and annotated query to measure token-level overlap, which can be computed by 
\begin{align}
    \text{precision} &= \frac{\#\text{common tokens}}{\#\text{tokens in predicted query}},\nonumber\\
    \text{recall} &= \frac{\#\text{common tokens}}{\#\text{tokens in golden query}},\\
    \text{Query F1} &= \frac{2\times\text{precision}\times\text{recall}}{\text{precision}+\text{recall}},\nonumber
\end{align}
where \#common tokens is the number of tokens that occur in both prediction and golden query.
\verb|Accuracy| is calculate by
\begin{align}
    \text{\#detect}_{explicit} &= \sum_{i=1}^N \mathbb{1}(ks_{pred}=explicit\:\&\:ks_{gold}=explicit),\nonumber\\
    \text{\#detect}_{implicit} &= \sum_{i=1}^N \mathbb{1}(ks_{pred}=implicit\:\&\:ks_{gold}=implicit),\\
    \text{Accuracy} &= \frac{\text{\#detect}_{explicit} + \text{\#detect}_{implicit} }{N} \times 100,\nonumber
\end{align}
where $\mathbb{1}(\cdot)$ is the indicator function, $N$ is the total number of dialogs containing inserted QA turns in the test set, $ks_{pred}$ and $ks_{gold}$ represent predicted and annotated knowledge sources, respectively. \#detect$_{explicit}$ (or \#detect$_{implicit}$) is the number of correctly predicted states indicating to use explicit (or implicit) external knowledge source.
\verb|Success Rate| is formulated as
\begin{equation}
    \frac{\sum_{i=1}^N\mathbb{1}(\frac{1}{n_{i}}\sum_{j=1}^{n_{i}} \text{Query-F1}(q_{pred}, q_{gold}) \ge 20)}{N},
\end{equation}
where $n_{i}$ is the number of inserted QA turns in $i$-th dialog, $q_{pred}$ and $q_{gold}$ represent predicted query and annotated search query, respectively. A dialog with inserted QA turns is considered successful in the QA task if the average \verb|Query F1| within the dialog is no less than 20.

In the full task setting, we report \verb|BLEU|, \verb|Inform|, \verb|Success| and \verb|Combine|. \verb|BLEU| score is computed on all TOD and QA responses. The computation of \verb|Inform| and \verb|Success| is different from that in the TOD setting. A dialog is considered successful in the full task if all the requested information is answered and the average query F1 of inserted QA turns is no less than 20. In the implementation, we use \verb|Success| in the QA task (average \verb|Query F1|$\ge$20) as an indicator. We only measure \verb|Success| and \verb|Inform| on dialogs whose average \verb|Query F1| of QA turns is no less than 20.
% In the implementation, we use \verb|Success| in the QA task (average \verb|Query F1|$\ge$20) as an indicator and only consider \verb|Success| for qualified dialogs. Similarly, we only compute \verb|Inform| for dialogs with an average query F1 greater than or equal to 20.

\def\answerable{\texttt{Answerable}}
\def\unanswerable{\texttt{Unanswerable}}

We conduct experiments on \multiwozqa{}, and further divide it into two subsets \answerable{} and \unanswerable{} based on the annotated question type (answerable or unanswerable) of inserted knowledge-seeking turns for ablations. Suppose a dialog contains both answerable and unanswerable questions. In that case, it is expanded to two examples, e.g., one containing TOD turns and answerable QA turns, and one consisting of TOD turns and unanswerable QA turns. Statistics of two subsets are shown in Table~\ref{tab:subsetstats}. \answerable{} includes 647 QA turns and 3981 TOD turns in the test set. The test set of \unanswerable{} contains 269 QA turns and 1888 TOD turns.
\begin{table}[htp]
  \caption{Statistics of \multiwozqa{} test set on the number of \answerable{} and \unanswerable{} questions.}
  \label{tab:subsetstats}
  \resizebox{0.55\columnwidth}{!}{
  \begin{tabular}{c|ccc}
    \toprule
     &\# TOD turns &\# QA turns &\# Dialogs\\
    \midrule
    \answerable{}  & 3981 & 647 & 577\\
    \unanswerable{} & 1888 & 269  & 259\\
    \bottomrule
  \end{tabular}}
\end{table}

\subsubsection{Models}
We train five models in an end-to-end manner and evaluate them in experiments. Training settings for models are summarized in Table~\ref{tab:models}.

\textbf{\baseline} is the closed-book baseline, which can only handle TOD (\texttt{T}) modeling and is unable to access external knowledge. It is trained on TOD turns in \multiwozqa{} and not exposed to the QA task in the training process.

\textbf{\closedbook{} } is the model with both TOD (\texttt{T}) and QA (\texttt{Q}) skills but not exposed to external knowledge sources during training and still in the closed-book setting. It only makes use of the dialog histories of QA turns during training. Search queries and external knowledge for QA turns are hidden. Compared to \baseline{}, it has "memorized" some knowledge in training and can utilize "memorized" information to answer questions.

\textbf{\explicit{}} is the model with TOD (\texttt{T}) modeling skill and able to handle QA (\texttt{Q}) with access to explicit external knowledge (\textit{w}/ EK). It is trained on full \multiwozqa. We utilize dialog histories, search queries, and selected knowledge for answerable QA turns and only use dialog histories for unanswerable QA turns. Queries and external knowledge for unanswerable questions are not provided during training. Therefore the model is expected to utilize explicit knowledge appropriately and answer questions requiring implicit information using memorized knowledge.

\textbf{\implicitres{}} is the proposed model that can handle TOD and QA tasks. It can access explicit knowledge from the Web and implicit knowledge using GPT-3 as a policy model (\texttt{GPT3PM}). It is trained on full \multiwozqa. We use dialog histories, search queries, and external knowledge for all QA turns in training. The augmented implicit knowledge is GPT-3 generated responses based on dialog histories for unanswerable questions. This model is anticipated to have the capability to consult explicit or implicit knowledge as needed in inference. When the model predicts to consult an implicit knowledge source, it prompts GPT-3 by dialog history and uses the generated response as knowledge. 

\textbf{\implicitk{}} is the proposed model that can handle TOD and QA tasks. It can acquire explicit knowledge from the Web and implicit knowledge using GPT-3 as a knowledge base (\texttt{GPT3KB}). It is trained on full \multiwozqa{} with a similar setting to \implicitres. The only difference is that the augmented implicit knowledge is GPT-3 generated knowledge snippets based on search queries. The model prompts GPT-3 using the predicted query.

\begin{table}[htp]
  \caption{Training settings of models. Mark $\cmark$ represents data available during training, and mark $\xmark$ denotes the information not provided. \baseline{} represents the model only with TOD (\texttt{T}) modeling skill while \closedbook{}  denotes the model with both TOD (\texttt{T}) and QA (\texttt{Q}) skills. \explicit{} represents the model with TOD and QA skills and the ability to access explicit knowledge (\textit{w}/ EK). \implicitres{} (\implicitk{}) indicates that the model uses GPT-3 as a policy model (knowledge base) to obtain implicit knowledge.}
  \label{tab:models}
  \resizebox{0.8\columnwidth}{!}{
  \begin{tabular}{c|c|c|c|c}
    \toprule
    \multirow{2}{*}{Model} & \multirow{2}{*}{TOD turns} & \multirow{2}{*}{QA turns} &\multicolumn{2}{c}{External knowledge source}\\
    \cline{4-5}
    & & & Explicit & Implicit\\
    \midrule
    \baseline & $\cmark$ & $\xmark$ & \textemdash & \textemdash\\
    \closedbook & $\cmark$ & $\cmark$ & \textemdash & \textemdash\\
    \explicit & $\cmark$ & $\cmark$ & Bing & \textemdash\\
    \implicitres & $\cmark$ & $\cmark$ & Bing & GPT-3 as policy model\\
    \implicitk & $\cmark$ & $\cmark$ & Bing & GPT-3 as knowledge base \\
    \bottomrule
  \end{tabular}}
\end{table}

\subsubsection{Implementation Details}
The implementation of models is based on Huggingface Pytorch Transformer \cite{Wolf20} T5-base model. Training examples are truncated with (or padded to) length of 512. To make sure input stings contain both dialog history and retrieved knowledge, we truncate dialog history on the left with a maximal lengths of 256. The maximal length of generation is 80 based on statistics in Table~\ref{tab:qastatistics}. The history window size is set to 2 and therefore dialog history $\boldsymbol{h} = \{ \boldsymbol{u}_{t-2},\boldsymbol{r}_{t-2}, \boldsymbol{u}_{t-1},\boldsymbol{r}_{t-1}, \boldsymbol{u}_{t}\}$ contains five utterances. We use AdamW optimizer \cite{Loshchilov18} with constant learning rate 0.001. Models are trained with a mini-batch of 6 on 4 Nvidia Tesla K80 until no decrease in validation loss is observed or up to 20 epochs.

\subsection{Main Results}
\begin{table}[htp]
  \caption{End-to-End evaluation of the full task on \multiwozqa}
  \label{tab:overall}
  \resizebox{0.57\columnwidth}{!}{
  \begin{tabular}{c|cccc}
    \toprule
    \multirow{2}{*}{Model} & \multicolumn{4}{c}{Full Task Evaluation} \\
    &BLEU & Success & Inform & Combined\\
    \midrule
    % \baseline & 14.06 & 2.83 & 5.16 & 18.05\\
    \baseline & 12.95 & 5.24 & 8.82 & 19.98\\
    % \closedbook & 14.28 & 5.90 & 9.34 & 21.90\\
    \closedbook & \textbf{14.99} & 5.63 & 9.97 & 22.79\\
    % \explicit & \textbf{14.81} & 28.38 & 42.01 & 50.01\\
    \explicit & 14.72 & 26.99 & 39.38 & 47.91\\
    % \implicitres & 14.49 & 30.71 & 49.63 & 54.66\\
    \midrule
    \implicitres & 13.52 & 28.90 & 45.01 & 50.47\\
    % \implicitk & 14.24 & \textbf{35.25} & \textbf{55.53} & \textbf{59.63}\\
    \implicitk & 14.18 & \textbf{31.33} & \textbf{52.17} & \textbf{55.93}\\
    \bottomrule
  \end{tabular}}
\end{table}
Table~\ref{tab:overall} shows the main experiment results on \multiwozqa. It is observed that \model{} models perform significantly better in \task{} task than other models in terms of the Combined score. With access to explicit Web knowledge, \explicit{} can complete both TOD and QA, indicating the necessity of incorporating external knowledge for \task. By leveraging both explicit and implicit knowledge, \implicitk{} obtains the best Combined score showing the effectiveness of the proposed model and the incorporation of different knowledge.

To have a better understanding of the variation in performance, we report the evaluation results of single tasks in Table~\ref{tab:TOD+qa}. The left column contains the evaluation results of TOD modeling. We notice that there is no obvious difference in TOD completion among models. The biggest difference in the Combined score in the TOD task is 6 points. However, \implicitk{} exceeds \baseline{} by about 36 points in the full task. This observation suggests that a better ability to handle the QA task leveraging external knowledge makes \model{} distinguished in the full task.
\begin{table}[htp]
  \caption{End-to-End evaluation of single tasks on \multiwozqa.}
  \label{tab:TOD+qa}
  \resizebox{\columnwidth}{!}{
  \begin{tabular}{c|cccc|cccc}
    \toprule
    \multirow{2}{*}{Model} & \multicolumn{4}{c|}{TOD Evaluation} & \multicolumn{4}{c}{QA Evaluation}\\
    & BLEU & Success & Inform & Combined & Accuracy & Success Rate & Query F1 & BLEU\\
    \midrule
    % \baseline & 15.51 & 36.49 & 53.44 & 60.48 & 0.00 & 11.67 & 7.87 & 2.78\\
    \baseline & 14.43 & 36.96 & 55.50 & 60.66 & 0.00 & 16.62 & 10.14 & 2.28\\
    % \closedbook & 14.71 & 35.26 & 55.04 & 59.86 & 0.00 & 16.95 & 9.47 & \textbf{6.36}\\
    \closedbook & \textbf{15.56} & \textbf{37.34} & \textbf{56.91} & \textbf{62.69} & 0.00 & 16.88 & 10.19 & 6.01\\
    % \explicit & 15.26 & \textbf{37.22} & 54.91 & 61.32 & 70.63 & 75.18 & 48.51 & 6.08\\
    \explicit & 15.25 & 36.45 & 54.22 & 60.59 & 70.63 & 72.25 & 47.91 & 5.99 \\
    % \implicitres & \textbf{15.52} & 32.31 & 52.83 & 58.09 & \textbf{98.25} & 93.12 & \textbf{61.58} & 5.74\\
    \midrule
    \implicitres & 14.49 & 32.86 & 51.53 & 56.69 & 86.90 & 85.81 & 56.07 & \textbf{6.22}\\
    % \implicitk & 15.20 & 36.98 & \textbf{58.35} & \textbf{62.86} & 96.94 & \textbf{93.61} & 60.33 & 6.20\\
    \implicitk & 15.03 & 34.02 & 56.14 & 60.11 & \textbf{92.03} & \textbf{90.15} & \textbf{59.24} & 6.01 \\
    \bottomrule
  \end{tabular}}
\end{table}

The right section of Table~\ref{tab:TOD+qa} shows the evaluation results of the QA task. Though \closedbook{} can generate more natural responses to questions than \baseline, both of them lack the ability to select appropriate external knowledge sources. They can only query the pre-defined database that does not contain the external information required by users and fail in the QA task. Though \explicit{} is far beyond models that cannot access external knowledge sources, there are certain disparities in comparing with \model. \implicitk{} has better performance in selecting appropriate knowledge sources and predicting precise search queries, while \implicitres{} can generate more fluent answers. The performance gap between \explicit{} and \model{} implies that incorporating explicit and implicit external knowledge sources benefits the QA task.

% Though they fail to predict correct knowledge sources, their predicted queries may overlap with golden search queries, making some QA turns successful. Comparing models with the ability to access external knowledge sources, \implicitk{} can select appropriate knowledge sources more accurately and generates more precise search queries, while \implicitres{} can generate more fluent answers. Though \explicit{} is far beyond models that cannot access external knowledge sources, there are certain disparities in comparing with models with the ability to consult implicit knowledge sources, implying the capability of consulting both explicit and implicit external knowledge sources may benefit the QA task.

\subsection{Evaluation on \answerable{} Questions}
\begin{table}[htp]
  \caption{End-to-End evaluation results on \answerable{} questions to demonstrate the importance of explicit external knowledge.}
  \label{tab:answerable}
  \resizebox{0.57\columnwidth}{!}{
  \begin{tabular}{c|cccc}
    \toprule
    \multirow{2}{*}{Model} & \multicolumn{4}{c}{Full Task Evaluation} \\
    &BLEU & Success & Inform & Combined\\
    \midrule
    % \baseline & 14.33 & 1.91 & 3.47 & 17.02\\
    \baseline & 13.16 & 4.40 & 6.97 & 18.84\\
    % \closedbook & 14.66 & 5.89 & 8.84 & 22.02\\
    \closedbook & \textbf{15.28} & 6.24 & 10.46 & 23.63\\
    % \explicit & \textbf{15.26} & 36.40 & 54.42 & 60.67\\
    \explicit & 15.15 & \textbf{36.70} & 53.40 & \textbf{60.20}\\
    \midrule
    % \implicitres & 15.01 & 31.72 & 52.00 & 56.87\\
    \implicitres & 14.13 & 33.39 & 51.38 & 56.52\\
    % \implicitk & 14.72 & \textbf{37.61} & \textbf{59.10} & \textbf{63.08}\\
    \implicitk & 14.77 & 33.94 & \textbf{56.15} & 59.81\\
    \bottomrule
  \end{tabular}}
\end{table}
Table~\ref{tab:answerable} contains evaluation results of the full task on \answerable. As what we have seen in Table~\ref{tab:overall}, \explicit{} and \model{} can complete the full task on answerable questions leveraging explicit knowledge from the Web. Without access to external knowledge sources, \baseline{} and \closedbook{} struggle to get good performance in the full task on answerable questions. The variation in performance indicates the importance of explicit knowledge for \task{}. We notice that there is a slight improvement of \closedbook{} compared to \baseline{}, which implies that the knowledge \closedbook{} has memorized during training can help the full task a little.

Table~\ref{tab:TOD+qa_on_ans} shows evaluation results of single tasks on \answerable{} questions. All models can complete the TOD task, and there is no huge difference in performance in TOD modeling.
However, when it comes to the QA task on answerable questions, \baseline{} and \closedbook{} are not satisfactory without access to external knowledge. Using knowledge memorized in the training process, \closedbook{} can generate more natural responses than \baseline{} but cannot compete with models with the ability to obtain explicit external knowledge. Able to acquire explicit knowledge from the Web, \explicit{} shows strong performance in handling answerable questions, which suggests that leveraging external knowledge is the key to success in the QA task. \model{} models obtain better performance in selecting appropriate knowledge sources and generating natural responses, implying that incorporating implicit knowledge can benefit the ability to utilize explicit knowledge.
\begin{table}[htp]
  \caption{End-to-End evaluation results of single tasks on \answerable{} questions to demonstrate the importance of explicit external knowledge for the QA task seeking external information.}
  \label{tab:TOD+qa_on_ans}
  \resizebox{\columnwidth}{!}{
  \begin{tabular}{c|cccc|cccc}
    \toprule
    \multirow{2}{*}{Model} & \multicolumn{4}{c|}{TOD Evaluation} & \multicolumn{4}{c}{QA Evaluation}\\
    & BLEU & Success & Inform & Combined & Accuracy & Success Rate & Query F1 & BLEU\\
    \midrule
    % \baseline & \textbf{15.77} & 37.95 & 55.29 & 62.39 & 0.00 & 8.32 & 6.53 & 2.23\\
    \baseline & 14.70 & \textbf{38.72} & 57.61 & 62.86 & 0.00 & 12.11 & 8.51 & 1.54\\
    % \closedbook & 14.83 & 36.40 & 57.37 & 61.71 & 0.00 & 14.04 & 8.16 & 7.59\\
    \closedbook & \textbf{15.54} & 37.80 & \textbf{58.17} & \textbf{63.52} & 0.00 & 15.96 & 9.56 & 7.29\\
    % \explicit & 15.45 & 37.61 & 55.81 & 62.16 & \textbf{100.00} & 97.05 & 63.44 & \textbf{8.06}\\
    \explicit & 15.32 & 37.80 & 55.05 & 61.74 & \textbf{100.00} & 97.25 & \textbf{63.94} & 7.91\\
    % \implicitres & 15.76 & 32.41 & 53.55 & 58.74 & \textbf{100.00} & 96.88 & 63.60 & 7.53\\
    \midrule
    \implicitres & 14.77 & 34.13 & 52.48 & 58.08 & \textbf{100.00} & 97.43 & 63.55 & \textbf{8.73}\\
    % \implicitk & 15.45 & \textbf{38.30} & \textbf{60.31} & \textbf{64.75} & \textbf{100.00} & \textbf{97.75} & \textbf{63.82} & 7.73\\
    \implicitk & 15.23 & 35.60 & 57.98 & 62.02 & \textbf{100.00} & \textbf{97.61} & 63.35 & 8.40\\
    \bottomrule
  \end{tabular}}
\end{table}

\subsection{Evaluation on \unanswerable{} Questions}
After realizing that explicit external knowledge is important for responding to answerable questions, we then examine whether only being able to access explicit knowledge can guarantee success in the QA task.
\begin{table}[htp]
  \caption{End-to-End evaluation results on \unanswerable{} questions to demonstrate the importance of implicit external knowledge.}
  \label{tab:unanserable}
  \resizebox{0.57\columnwidth}{!}{
  \begin{tabular}{c|cccc}
    \toprule
    \multirow{2}{*}{Model} & \multicolumn{4}{c}{Full Task Evaluation} \\
    &BLEU & Success & Inform & Combined\\
    \midrule
    % \baseline & 13.75 & 4.63 & 8.49 & 20.31\\
    \baseline & 12.75 & 6.95 & 12.74 & 22.59\\
    % \closedbook & 13.71 & 5.79 & 10.42 & 21.82\\
    \closedbook & \textbf{14.44} & 4.25 & 8.49 & 20.81\\
    % \explicit & \textbf{14.06} & 8.50 & 11.58 & 24.10\\
    \explicit & 14.16 & 4.25 & 6.56 & 19.56\\
    % \implicitres & 13.66 & 28.18 & 44.01 & 49.75\\
    \midrule
    \implicitres & 12.47 & 18.53 & 30.50 & 36.98\\
    % \implicitk & 13.49 & \textbf{30.51} & \textbf{47.49} & \textbf{52.49}\\
    \implicitk & 13.33 & \textbf{25.10} & \textbf{42.47} & \textbf{47.11}\\
    \bottomrule
  \end{tabular}}
\end{table}

Evaluation results on \unanswerable{} are listed in Table~\ref{tab:unanserable}. With access to both explicit and implicit knowledge, \model{} models overperform the other models by over 17 points in the Combined score. We have observed that \explicit{} has good performance in responding to answerable questions leveraging explicit Web knowledge. However, it suffers in the evaluation of handling unanswerable questions, which indicates that explicit knowledge does not help answer unanswerable questions.

We also conduct experiments to evaluate models on single tasks and report results in Table~\ref{tab:TOD+qa_on_unans}. \closedbook{} performs better in TOD modeling but does not exceed other models much. The disparities become obvious in the QA task. \model{} models can access the implicit knowledge source as needed, making them succeed in more QA turns. In contrast, without the ability to consult the implicit knowledge source, the other three models tend to answer questions by querying the pre-defined database and fail to provide users with useful information. The huge lead of \model{} shows that the strategy to incorporate implicit and explicit external knowledge is efficient.
\begin{table}[htp]
  \caption{End-to-End evaluation results of single tasks on \unanswerable{} questions to demonstrate the importance of implicit external knowledge for the QA task.}
  \label{tab:TOD+qa_on_unans}
  \resizebox{\columnwidth}{!}{
  \begin{tabular}{c|cccc|cccc}
    \toprule
    \multirow{2}{*}{Model} & \multicolumn{4}{c|}{TOD Evaluation} & \multicolumn{4}{c}{QA Evaluation}\\
    & BLEU & Success & Inform & Combined & Accuracy & Success Rate & Query F1 & BLEU\\
    \midrule
    % \baseline & 15.14 & 33.59 & 49.42 & 56.65 & 0.00 & 18.92 & 11.09 & 3.61\\
    \baseline & 14.01 & 33.98 & 51.35 & 56.67 & 0.00 & 25.87 & 14.07 & 3.30\\
    % \closedbook & 14.71 & 32.82 & 50.58 & 56.41 & 0.00 & 22.78 & 12.61 & \textbf{4.28}\\
    \closedbook & \textbf{15.64} & \textbf{36.29} & \textbf{53.67} & \textbf{60.62} & 0.00 & 18.53 & 11.72 & \textbf{3.81}\\
    % \explicit & 15.07 & \textbf{37.07} & 53.67 & 60.44 & 0.00 & 21.62 & 12.60 & 2.69\\
    \explicit & 15.41 & 32.82 & 51.35 & 57.50 & 0.00 & 15.06 & 9.36 & 2.64\\
    % \implicitres & \textbf{15.19} & 32.43 & 51.74 & 57.27 & \textbf{94.05} & \textbf{83.78} & \textbf{56.70} & 2.79\\
    \midrule
    \implicitres & 14.07 & 30.50 & 49.81 & 54.23 & 55.39 & 58.69 & 38.07 & 2.04\\
    % \implicitk & 14.89 & 35.52 & \textbf{55.60} & \textbf{60.45} & 89.59 & 82.63 & 51.93 & 3.77\\
    \implicitk & 14.97 & 30.89 & 52.12 & 56.47 & \textbf{72.86} & \textbf{72.20} & \textbf{49.35} & 2.09\\
    \bottomrule
  \end{tabular}}
\end{table}

\subsection{Is implicit external knowledge enough for QA?}
The previous analysis suggests that implicit external knowledge is pivotal for the success of QA and full tasks. Next, we investigate whether implicit knowledge only can eliminate the need for explicit knowledge.
%The previous analysis suggests that it is impossible to handle the QA task with access only to the explicit external knowledge and implicit external knowledge is pivotal for the success of QA and full tasks. Next, we investigate whether implicit knowledge only can eliminate the need for explicit knowledge.

Table~\ref{tab:all_qa} shows the evaluation results of GPT-3 and \model{} models on all QA turns. We use GPT-3 as a control since it is proven to have strong abilities in various NLP tasks, including question answering \cite{Brown20} and knowledge retrieval \cite{Li20}. We also use GPT-3 generation as implicit knowledge in the training procedure. Therefore we consider GPT-3 contains rich implicit knowledge and compare it with \model{} models that are capable of utilizing both explicit and implicit knowledge. Though GPT-3 performs better in responding to unanswerable questions, it falls behind much in answerable questions, indicating the importance of explicit knowledge in the QA task. Our strategy to combine explicit with implicit knowledge does benefit the QA task.
\begin{table}[htp]
  \caption{Evaluation results on all QA turns to demonstrate the necessity to incorporate explicit and implicit knowledge.}
  \label{tab:all_qa}
  \resizebox{0.75\columnwidth}{!}{
  \begin{tabular}{c|ccc}
    \toprule
    Model & Answerable BLEU & Unanswerable BLEU& Overall BLEU\\
    \midrule
    GPT-3 & 2.20 & \textbf{3.73} & 2.80\\
    \midrule
    % \implicitres & 7.53 & 2.79 & 5.74 \\
    \implicitres & \textbf{8.73} & 2.04 & \textbf{6.22}\\
    % \implicitk & \textbf{7.73} & \textbf{3.77} & \textbf{6.20}\\
    \implicitk & 8.40 & 2.09 & 6.01\\
  \bottomrule
\end{tabular}}
\end{table}

\subsection{Human Evaluation}

We conduct a turn-level pairwise t-test for human evaluation to assess \model{}'s performance. \explicit{} and \baseline{} are compared to demonstrate the importance of explicit external knowledge, and \explicit{} and \implicitk{} are to evaluate the effectiveness of incorporating implicit knowledge.

We randomly sample 716 responses for each model. For each example, we hire workers with lifetime HIT acceptance \% > 95 on Amazon Mechanical Turk and present workers with dialog history and two responses from different models. We consider three questions rated on a 5-point Likert scale, : 1) Usefulness measures how well a response can provide expected information. 2) Humanness measures how fluent a response is and whether it is coherent with the dialog context. 3) Safety measures whether a response is socially safe. A safe response is considered not to contain toxic, biased, or misleading content.

The results of human evaluation are reported in Table~\ref{tab:human}. 5-point Likert is converted into a Win/Tie /Loss scale.
The left column contains the comparison between \baseline{} and \explicit{}. We report the percentage that \explicit{} wins, ties with, and loses to \baseline{}. The model with access to explicit knowledge beats \baseline{} on all the three aspects, showing that leveraging explicit knowledge helps to generate more informative, natural, and human-like responses. The right column shows the results of the comparison of \explicit{} and \implicitk{}, including the probability that \implicitk wins, ties with, and loses to \explicit{}. Our proposed model generates responses with significantly better informativeness and fluency. The success of \implicitk{} indicates that our strategy to utilize both explicit and implicit knowledge is beneficial to designing more human-like dialog systems.
\begin{table}[htp]
  \caption{Human evaluation results. All rows are significant ($p<0.05$ with paired t-test) except the last. In the left section, "Win" (or "Loss") stands for the percentage that \explicit{} wins (or loses). In the right section, "Win" (or "Loss") represents the percentage that \implicitk{} wins (or loses).}
  \label{tab:human}
  \resizebox{0.95\columnwidth}{!}{
  \begin{tabular}{p{2cm}|p{1.5cm}p{1.5cm}p{1.5cm}|p{1.5cm}p{1.5cm}p{1.5cm}}
    \toprule
    & \multicolumn{3}{c|}{\explicit{} vs. \baseline{}} & \multicolumn{3}{c}{\implicitk{} vs. \explicit{}}\\
    \cline{2-7}
    & \hfil Win & \hfil Tie & \hfil Loss &\hfil Win &\hfil Tie &\hfil Loss \\
    \midrule
    \hfil Usefulness &\hfil \textbf{34.5} &\hfil 37.9 &\hfil 27.6 &\hfil \textbf{32.3} &\hfil 41.4 &\hfil 26.3\\
    \hfil Humanness & \hfil \textbf{39.2} &\hfil 32.8 & \hfil 28.0 &\hfil \textbf{38.4} &\hfil 38.4 &\hfil 23.2\\
    \hfil Safety &\hfil \textbf{31.9} &\hfil 41.8 &\hfil 26.3 &\hfil \textbf{32.3} &\hfil 37.4 &\hfil 30.3\\
    \bottomrule
    \end{tabular}}
\end{table}

\subsection{Case Study}
We show example dialog segments and corresponding output results of models $(\boldsymbol{s}, \boldsymbol{r})$ in Table~\ref{tab:casestudy}. Without access to external knowledge sources, \closedbook{} predicts the knowledge source to be the pre-defined database by mistake in both examples and fails to provide expected information to the user. In the first example, \explicit{} and \implicitk{} can acquire knowledge from the Web and respond to the user with accurate information. In the second example, \explicit{} fails to consult implicit knowledge sources and gives incorrect information to the user. In contrast, \implicitk{} succeeds in providing correct information to users. However, we can notice that \model{} cannot provide as many details as humans. In the second example, the response of \implicitk{} only suggests that the taxi booking can be canceled, while the human response also includes restrictions on cancellation. This indicates that there is still room to improve \model{} to generate correct and detailed responses.
\begin{table}[htp]
  \caption{Comparison of state prediction and response generation results when models occur inserted QA turns in \multiwozqa. "Human" denotes ground truth results annotated by workers.}
  \label{tab:casestudy}
  \resizebox{\columnwidth}{!}{
  \begin{tabular}{l|l}
    \toprule
    \multicolumn{2}{l}{\textbf{Example 1}} \\
    \multicolumn{2}{l}{U: I need a train leaving on Friday. } \\
    \multicolumn{2}{l}{R: Where are you departing from and what is your destination?}\\
    \multicolumn{2}{l}{U: The train should depart from Cambridge and should go to Ely.}\\
    \multicolumn{2}{l}{R: Is there a particular time you'd like to leave?}\\
    \midrule
    \multicolumn{2}{l}{U: My itinerary isn't confirmed yet, so what is the cancellation policy for the train?}\\
    \midrule
    \multirow{2}{*}{\closedbook} & $\boldsymbol{s}:$ Dataset: train destination = ely ; day = friday ; departure = cambridge \\
    & $\boldsymbol{r}:$ There are [value\_count] trains that meet your criteria.\\
    \midrule
    \multirow{2}{*}{\explicit} & $\boldsymbol{s}:$ Explicit: cancellation policy for train\\
    & $\boldsymbol{r}: $ You can cancel your train ticket booking by calling them or cancel your booking by\\ & going to the app where you booked it to cancel the booking.\\
    \midrule
    \multirow{2}{*}{\implicitk} & $\boldsymbol{s}: $ Explicit: cancellation policy for train\\
    & $\boldsymbol{r}:$ You can change or cancel your train ticket booking before the train departs.\\
    \midrule
    \multirow{2}{*}{Human} & $\boldsymbol{s}:$ Explicit: cambridge train cancellation policy\\
    & $\boldsymbol{r}:$ Don't worry, your ticket will be refunded. Any other questions?\\
    \midrule
    \midrule
    \multicolumn{2}{l}{\textbf{Example 2}} \\
    \multicolumn{2}{l}{......}\\
    \multicolumn{2}{l}{U: I want to book a taxi to commute between the two places.} \\
    \multicolumn{2}{l}{R: To which direction will you be going?}\\
    \multicolumn{2}{l}{U: Going from El Shaddai to the Cow Pizza Kitchen and Bar, and I need to arrive in time for my reservation}\\ \multicolumn{2}{l}{at the restaurant.}\\
    \multicolumn{2}{l}{R: Okay I will look into this for you and be back in a moment.}\\
    \midrule
    \multicolumn{2}{l}{U: Will I be able to cancel my taxi booking if my plans change later on?}\\
    \midrule
    \multirow{2}{*}{\closedbook} & $\boldsymbol{s}:$ Dataset: taxi destination = the cow pizza kitchen and bar ; departure = el shaddai \\
    & $\boldsymbol{r}:$ Yes, I have booked you a taxi. The contact number is [taxi\_phone].\\
    \midrule
    \multirow{2}{*}{\explicit} & $\boldsymbol{s}:$ Dataset: taxi destination = the cow pizza kitchen and bar ; departure = el shaddai \\
    & $\boldsymbol{r}: $ I am sorry, but I am unable to cancel your booking. Would you like to try a different\\& taxi booking?\\
    \midrule
    \multirow{2}{*}{\implicitk} & $\boldsymbol{s}: $ Implicit: can I cancel my taxi booking later on\\
    & $\boldsymbol{r}:$ Yes, I will cancel your taxi booking if you change your mind later on.\\
    \midrule
    \multirow{2}{*}{Human} & $\boldsymbol{s}:$ Implicit: el shaddai taxi cancel booking\\
    & $\boldsymbol{r}:$ You can cancel 24 hours in advance. After that there is a 10 gbp fee. Does this help\\& you?\\
    \bottomrule
  \end{tabular}}
\end{table}

\section{Conclusion and Future Work}

In this work, we define a new task \task{} that combines task-oriented dialog and information seeking via external knowledge (e.g., the Web and language models).
%QA with access to external knowledge and 
We construct a new dataset \multiwozqa{} with manually composed questions, search queries, and knowledge grounded responses. To tackle this task, we propose \model, a unified model that seamlessly incorporates QA capabilities seeking external information into an end-to-end task-oriented dialog agent. %with explicit and implicit knowledge. 
Experimental results on \multiwozqa{} indicate that by combining implicit and explicit knowledge, \model{} is able to handle the fused task and outperform closed-book baselines.
%TOD and QA grounded on external knowledge. 
We believe that the newly proposed  task \task{} may represent a significant step towards building human-like conversational AI agents, and \multiwozqa{} can help facilitate research in this direction.

For future work, we believe it is important to explore evaluation metrics that consider both humanness and factuality of dialogs. In \task{}, we have only considered adding QA capabilities to task-oriented dialog agents. However, in practice, humans have diverse information-seeking needs. As such, it is worthwhile to build a more comprehensive task involving multiple skills (e.g., recommendation and personalization), and benchmark the progress of human-like conversational AI agents. 
%We leave it as future work.

% [discussion of future work needed.]
% \begin{itemize}
%     \item factualness of responses
%     \item Better metrics for new task setting
%     \item more skills
% \end{itemize}

%%
%% The acknowledgments section is defined using the "acks" environment
%% (and NOT an unnumbered section). This ensures the proper
%% identification of the section in the article metadata, and the
%% consistent spelling of the heading.
\begin{acks}

\end{acks}

%%
%% The next two lines define the bibliography style to be used, and
%% the bibliography file.
\bibliographystyle{ACM-Reference-Format}
\bibliography{paper.bib}

% \printbibliography
%%
%% If your work has an appendix, this is the place to put it.
\end{document}